\documentclass[%
    paper=A4,               
    twoside=true,           
    openright,              
    parskip=full,           
    chapterprefix=true,     
    11pt,                   
    headings=normal,        
    bibliography=totoc,     
    listof=totoc,           
    titlepage=on,           
    captions=tableabove,    
    draft=false            
]{scrreprt}%

\PassOptionsToPackage{utf8}{inputenc}
\usepackage{inputenc}

\newcommand{\plainThesisTitle}{Automated Machine Learning for Multi-Label Classification}
\newcommand{\thesisTitle}{Automated Machine Learning for\\ Multi-Label Classification}
\newcommand{\thesisName}{Marcel Wever}
\newcommand{\thesisSubject}{Dissertation}
\newcommand{\thesisDate}{March 01, 2022}

\newcommand{\thesisFirstReviewer}{Prof. Dr. Eyke H\"{u}llermeier}
\newcommand{\thesisFirstReviewerUniversity}{\protect{Ludwig-Maximilians-Universität München}}
\newcommand{\thesisFirstReviewerDepartment}{K{\"u}nstliche Intelligenz und Maschinelles Lernen}

\newcommand{\thesisSecondReviewer}{Prof. Dr. Axel-Cyrille Ngonga Ngomo}
\newcommand{\thesisSecondReviewerUniversity}{\protect{Paderborn University}}
\newcommand{\thesisSecondReviewerDepartment}{Data Science}

\newcommand{\thesisThirdReviewer}{Prof. Dr. Bernd Bischl}
\newcommand{\thesisThirdReviewerUniversity}{\protect{Ludwig-Maximilians-Universität München}}
\newcommand{\thesisThirdReviewerDepartment}{Statistical Learning \& Data Science}

\newcommand{\thesisFirstSupervisor}{Prof. Dr. Eyke H\"{u}llermeier}

\newcommand{\thesisUniversity}{}
\newcommand{\thesisUniversityDepartment}{Department of Electrical Engineering,\\ Computer Science and Mathematics}

\newcommand{\thesisUniversityGroup}{Intelligent Systems and Machine Learning}
\newcommand{\thesisUniversityCity}{Paderborn}
\newcommand{\thesisUniversityStreetAddress}{Pohlweg 51}
\newcommand{\thesisUniversityPostalCode}{33098}

\usepackage[english]{babel} 
\PassOptionsToPackage{
    figuresep=colon,%
    sansserif=false,%
    hangfigurecaption=false,%
    hangsection=true,%
    hangsubsection=true,%
    colorize=reduced,%
    colortheme=upbisg,%
    bibsys=biber,%
    bibfile=bib-refs,%
    bibstyle=alphabetic,
    wrapfooter=false,%
}{cleanthesis}
\usepackage{cleanthesis}

\usepackage{mathtools}

\hypersetup{
    pdftitle={\plainThesisTitle},pdfsubject={\thesisSubject},pdfauthor={\thesisName},plainpages=false,colorlinks=false,pdfborder={0 0 0},breaklinks=true,bookmarksnumbered=true,bookmarksopen=true}

\usepackage{placeins}
\usepackage{subcaption}
\usepackage{array}
\usepackage{multirow}
\usepackage{graphicx}
\usepackage{booktabs}

\usepackage{amsmath}
\usepackage{amssymb}
\usepackage{nicefrac}
\usepackage{stmaryrd}
\usepackage{pifont}

\renewcommand{\vec}[1]{\boldsymbol{#1}}
\newcommand{\given}{\, | \,}
\newcommand{\defeq}{:=}
\newcommand{\fromto}{\longrightarrow}

\newcommand{\ceil}[1]{\lceil #1 \rceil}

\preto\fullcite{\AtNextCite{\defcounter{maxnames}{99}}}

\usepackage{pdfpages}
\usepackage{ifthen}
\newboolean{incpapers} 
\setboolean{incpapers}{false} 

\DeclareUnicodeCharacter{0301}{\'{e}}

\begin{document}

\renewcaptionname{english}{\figurename}{Figure}
\renewcaptionname{english}{\tablename}{Table}

\pagenumbering{roman}			
\pagestyle{empty}				
%
\begin{titlepage}
	\pdfbookmark[0]{Cover}{Cover}
	\flushright
	\hfill
	\vfill
	{\LARGE\thesisTitle \par}
	\rule[5pt]{\textwidth}{.4pt} \par
	{\Large\thesisName}
	\vfill
	\textit{\large\thesisDate} \\
\end{titlepage}

\begin{titlepage}
	\pdfbookmark[0]{Titlepage}{Titlepage}
	\tgherosfont
	
	\begin{figure}
	\begin{minipage}[t]{8.5cm}		
	\includegraphics[height=1.8cm]{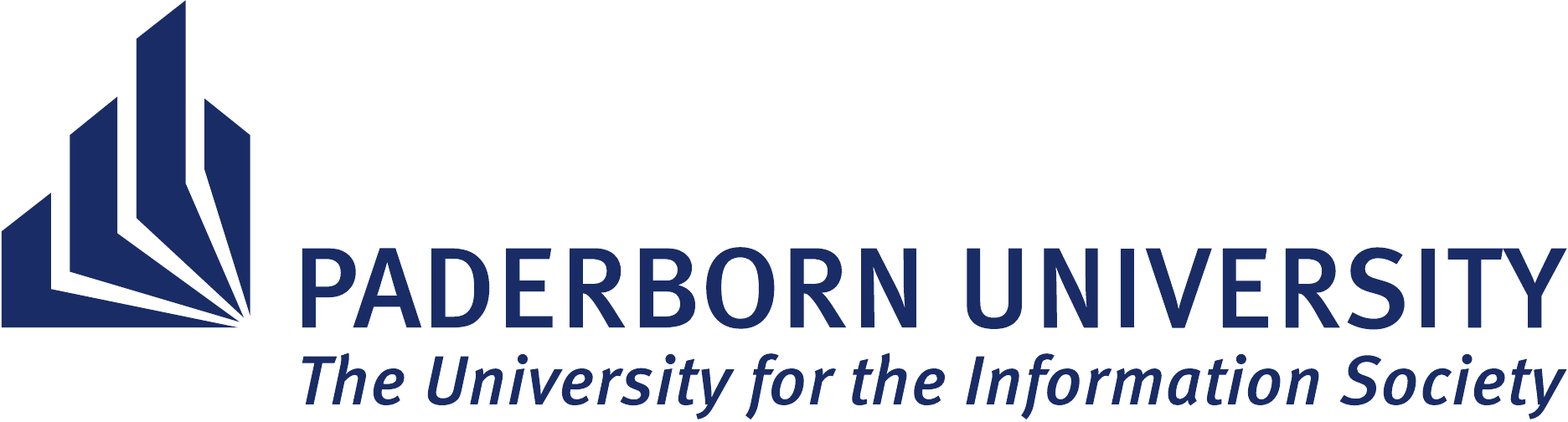}\\
	\textsf{\small{\hspace*{1.3cm}Department of Electrical Engineering,\\
	\hspace*{1.3cm}Computer Science and Mathematics\\
		\hspace*{1.3cm}Warburger Straße 100 \\
		\hspace*{1.3cm}33098 Paderborn
		}}
	\end{minipage}
	\hfill
	\begin{minipage}[t]{4.7cm}
	\includegraphics[height=1.8cm]{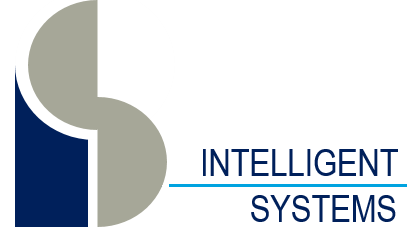}\\
	\textsf{
	\hspace*{0.1cm}\small{Intelligent Systems Group (ISG)}
	}
	\end{minipage}
	\end{figure}
	
	\centering

	\vfill
	{\large \thesisSubject} \\[5mm]
	In partial fulfillment of the requirements for the academic degree of\\
    \textbf{Doctor rerum naturalium (Dr. rer. nat.)}\\[5mm]
	{\LARGE \color{ctcolortitle}\textbf{\thesisTitle} \\[10mm]}
	{\Large \thesisName} \\

	\vfill
	\begin{minipage}[t]{.27\textwidth}
		\raggedleft
		\textit{1. Reviewer}
	\end{minipage}
	\hspace*{15pt}
	\begin{minipage}[t]{.65\textwidth}
		{\Large \thesisFirstReviewer} \\
	  	{\small \thesisFirstReviewerDepartment} \\[-1mm]
		{\small \thesisFirstReviewerUniversity}
	\end{minipage} \\[5mm]
	\begin{minipage}[t]{.27\textwidth}
		\raggedleft
		\textit{2. Reviewer}
	\end{minipage}
	\hspace*{15pt}
	\begin{minipage}[t]{.65\textwidth}
		{\Large \thesisSecondReviewer} \\
	  	{\small \thesisSecondReviewerDepartment} \\[-1mm]
		{\small \thesisSecondReviewerUniversity}
	\end{minipage} \\[5mm]
	\begin{minipage}[t]{.27\textwidth}
		\raggedleft
		\textit{3. Reviewer}
	\end{minipage}
	\hspace*{15pt}
	\begin{minipage}[t]{.65\textwidth}
		{\Large \thesisThirdReviewer} \\
	  	{\small \thesisThirdReviewerDepartment} \\[-1mm]
		{\small \thesisThirdReviewerUniversity}
	\end{minipage} \\[10mm]
	\begin{minipage}[t]{.27\textwidth}
		\raggedleft
		\textit{Supervisor}
	\end{minipage}
	\hspace*{15pt}
	\begin{minipage}[t]{.65\textwidth}
		\thesisFirstSupervisor
	\end{minipage} \\[10mm]

	\thesisDate \\

\end{titlepage}

\hfill
\vfill
{
	\small
	\textbf{\thesisName} \\
	\textit{\thesisTitle} \\
	\thesisSubject, \thesisDate \\
	Reviewers: \thesisFirstReviewer,\\ \hspace*{1.78cm}\thesisSecondReviewer,\\ \hspace*{1.78cm}\thesisThirdReviewer \\
	Supervisor: \thesisFirstSupervisor \\[1.5em]
	\textbf{\thesisUniversity} \\
	\textit{\thesisUniversityGroup} \\
	\thesisUniversityDepartment \\
	\thesisUniversityStreetAddress \\
	\thesisUniversityPostalCode\ \thesisUniversityCity
}
\cleardoublepage

\chapter*{Danksagung}
Es gibt viele Menschen, die auf ihre ganz eigene Art und Weise zu dieser Arbeit beigetragen haben, und ich möchte an dieser Stelle eben diesen Menschen meinen herzlichen Dank aussprechen.

Zuallererst möchte ich meinem Doktorvater Eyke danken, für die Möglichkeit der Promotion, sowie das Vertrauen und die Förderung, für die Inspiration und die Mitarbeit an meinen Projekten und Papieren - auch außerhalb der gängigen Öffnungszeiten.
Weiterer Dank gilt auch meiner Promotionskommission, bestehend aus Axel, Bernd, Henning und Matthias, für ihre Bereitschaft, sich mit meiner Arbeit kritisch auseinanderzusetzen und an meiner Disputation teilzunehmen.
Eyke, Axel und Bernd danke ich außerdem für die Anfertigung ihrer Gutachten.

Der ehemaligen Fachgruppe Intelligente Systeme und Maschinelles Lernen an der Universität Paderborn sowie der neuen Fachgruppe Künstliche Intelligenz und Maschinelles Lernen an der LMU möchte ich dafür Danke sagen, mich in ihre Reihen aufgenommen, begleitet und unterstützt zu haben. Ohne dieses Umfeld von Kolleg:innen wäre vieles von dem, was ich in dieser Zeit erreicht habe, nicht möglich gewesen. Besonders danken möchte ich an dieser Stelle Elisabeth für ihre unermüdliche Unterstützung bei allen nicht-wissenschaftlichen Belangen, meinem Bürokollegen Alexander für viele wertvolle Diskussionen und nochmals Alexander sowie Viktor für das Korrekturlesen dieser Arbeit. Ein weiteres Dankeschön möchte ich den studentischen Hilfskräften aussprechen, die mich bei meinen Arbeiten über die Jahre unterstützt haben.

Außerdem möchte ich mich noch bei einem weiteren Kollegen und Freund bedanken. Felix hat mich von Tag Null an mit ins Boot geholt, mir gezeigt, unter welchem Kurs sich die Segel am besten mit Wind füllen und wie man mit den unterschiedlichsten Wetterlagen umgeht. Danke dafür, dass ich auch dann und wann mal ins kalte Wasser geschmissen wurde, damit ich meine eigenen Erfahrungen machen konnte.

Während meiner Zeit als Doktorand hatte ich zudem das Privileg im DFG-Projekt "InterGramm" sowie im DFG Sonderforschungsbereich (SFB) 901 "On-The-Fly Computing" mitzuarbeiten und im Rahmen des SFBs diese Arbeit schreiben zu dürfen. An dieser Stelle möchte ich auch nochmal dem Geschäftsführer des SFBs, Ulf, für sein Engagement und Unterstützung bei allen erdenklichen Fragen danken.

Zu guter Letzt möchte ich meinen Eltern für ihre steten Bemühungen, mich von Beginn an zu bestärken und in allem zu fördern, von ganzem Herzen danken. Zusätzlich danke ich auch meiner Familie und meinen Freunden für ihre Unterstützung auf Schritt und Tritt, für all die wertvollen Erinnerungen und dafür, dass sie immer für mich da sind. Ganz besonders möchte ich meiner Freundin Katharina danken, die vor allem auch dann für mich da war, wenn es mal etwas stressiger und chaotischer wurde.

\pagestyle{plain}				
%
\pdfbookmark[0]{Zusammenfassung}{Zusammenfassung}

\chapter*{Zusammenfassung}\label{sec:abstract-de} 
\vspace*{-10mm}
Das Ziel des automatisierten maschinellen Lernens (AutoML) ist es, zugeschnitten auf einen gegebenen Datensatz, Algorithmen für das maschinelle Lernen (ML) zu wählen, zu konfigurieren und in Form von ML-Pipelines zu kombinieren.
Für überwachte Lernaufgaben, insbesondere binäre und multinomiale Klassifikation, auch als Single-Label-Klassifikation (engl.: single-label classification; SLC) bezeichnet, haben solche AutoML-Ansätze vielversprechende Ergebnisse geliefert.
Die Aufgabe der Multi-Label-Klassifikation (engl.: multi-label classification; MLC), bei der Datenpunkte mit einer Menge von Klassenlabels anstelle eines einzelnen Klassenlabels assoziiert werden, hat bisher deutlich weniger Aufmerksamkeit erhalten.
Im Kontext der Multi-Label-Klassifikation ist die datenspezifische Auswahl und Konfiguration von Multi-Label-Klassifikatoren selbst für Experten auf diesem Gebiet eine Herausforderung, da es sich um ein hochdimensionales Optimierungsproblem mit hierarchischen Abhängigkeiten über mehrere Ebenen hinweg handelt.
Während der Raum von ML-Pipelines für SLC bereits äußerst viele Kandidaten umfasst, übertrifft die Größe des MLC-Suchraums die des SLC-Suchraums um mehrere Größenordnungen.

Im ersten Teil dieser Arbeit wird ein neuartiger AutoML-Ansatz für Single-Label-Klassifikationsaufgaben entwickelt, der ML-Pipelines optimiert, die aus maximal zwei Algorithmen bestehen. Dieser Ansatz wird dann erweitert, um zunächst Pipelines von unbegrenzter Länge und schließlich die komplexen hierarchischen Strukturen von Multi-Label-Klassifikatoren zu konfigurieren.
Außerdem untersuchen wir, wie gut Ansätze, die den Stand der Technik im Bereich AutoML für Single-Label-Klassifikationsaufgaben bilden, mit der erhöhten Komplexität des AutoML Problems für Multi-Label-Klassifikation skalieren.

Im zweiten Teil wird untersucht, wie Methoden für SLC und MLC flexibler konfiguriert werden können, um die zur Verfügung stehenden Daten besser zu generalisieren, und wie die Effizienz von ausführungsbasierten AutoML-Systemen mit Hilfe von Laufzeitvorhersagen für ML-Pipelines gesteigert werden kann.

\pdfbookmark[0]{Abstract}{Abstract}
\chapter*{Abstract}\label{ch:abstract-en} 
Automated machine learning (AutoML) aims to select and configure machine learning algorithms and combine them into machine learning pipelines tailored to a dataset at hand.
For supervised learning tasks, most notably binary and multinomial classification, aka single-label classification (SLC), such AutoML approaches have shown promising results.
However, the task of multi-label classification (MLC), where data points are associated with a set of class labels instead of a single class label, has received much less attention so far.
In the context of multi-label classification, the data-specific selection and configuration of multi-label classifiers are challenging even for experts in the field, as it is a high-dimensional optimization problem with multi-level hierarchical dependencies.
While for SLC, the space of machine learning pipelines is already huge, the size of the MLC search space outnumbers the one of SLC by several orders.

In the first part of this thesis, we devise a novel AutoML approach for single-label classification tasks optimizing pipelines of machine learning algorithms, consisting of two algorithms at most.
This approach is then extended first to optimize pipelines of unlimited length and eventually configure the complex hierarchical structures of multi-label classification methods.
Furthermore, we investigate how well AutoML approaches that form the state of the art for single-label classification tasks scale with the increased problem complexity of AutoML for multi-label classification.

In the second part, we explore how methods for SLC and MLC could be configured more flexibly to achieve better generalization performance and how to increase the efficiency of execution-based AutoML systems.		
\cleardoublepage
%
%
\setcounter{tocdepth}{2}		
\tableofcontents				
\cleardoublepage

\pagenumbering{arabic}			
\setcounter{page}{1}			
\pagestyle{maincontentstyle} 	

%
\chapter{Introduction}\label{sec:intro}

Within the past decade, the demand for artificial intelligence and, in particular, machine learning (ML) functionality has grown exceedingly fast and is still growing steadily.
Certainly, this can be attributed to relevant and media-effective breakthroughs that have achieved superhuman performance:
In 2011, for example, IBM's Watson already beat two champions in Jeopardy \cite{ferrucci2012introduction,ferrucci2013watson}, machine facial recognition managed to achieve an accuracy of more than 97\% in 2014 \cite{taigman2014deepface}, and in 2016 AlphaGo was the first computer program to beat professional human Go players \cite{silver2016mastering,silver2017mastering}.
Beyond that, machine learning applications can be found in more and more parts of society and the economy \cite{rudin2014machine,jordan2015machine,stateOfAI}.

However, the engineering of such applications is a non-trivial endeavor and requires expertise in machine learning that end users typically do not have.
More specifically, there exists a plethora of different machine learning algorithms which work differently well, depending on the tasks and data.
Additionally, most of these algorithms expose parameters, so-called hyper-parameters\footnote{Hyper-parameters are parameters to control the \textit{learning process}, e.g., the number of neurons in a neural network. In turn, the learning process induces the parameters of a model from data, e.g., the weights of a neural network. For example, the learning rate is a hyper-parameter of the learning process for neural networks.}, which need to be tuned to the data at hand to achieve the best possible performance \cite{libre,rivolli2020empirical}.
To bridge the gap between supply and demand, the field of automated machine learning (AutoML) emerged to help non-experts access machine learning technology on the one hand and, on the other hand, to support machine learning experts in their work relieving them of some of the tedious tasks.

AutoML is the vision of automating as much of the data science process as possible, starting from raw data and providing a complete pipeline of machine learning algorithms.
Such a pipeline may comprise algorithms for pre-processing the (raw) data, constructing or selecting features, and ultimately learning a model.
The problem of AutoML was first formally stated in \cite{autoweka} as the combined algorithm selection and hyper-parameter optimization (CASH) problem, optimizing for the generalization performance of the respective algorithm choice and its configuration.
Provided a CASH problem specification in terms of (i) a model of the search space, describing the space of available machine learning pipelines, and (ii) an evaluation function to assess the quality of solution candidates, the problem is usually treated as a sampling-based black-box optimization problem.
More specifically, an optimization algorithm is employed to traverse the search space in a trial-and-error fashion by executing solution candidates, i.e., machine learning pipelines, on the particular data.
First approaches to AutoML are based on random search \cite{DBLP:conf/nips/BergstraBBK11}, Bayesian optimization \cite{autoweka,autoweka2,autosklearn}, or genetic programming \cite{tpot,recipe}.

While these AutoML systems show promising performances for regression and binary or multi-class classification tasks, in the following referred to as single-label classification (SLC), where each data point is associated with a numeric value or a class label respectively, the task of multi-label classification (MLC) received far less attention.
In MLC, instead of only a single class label, each data point is associated with a (sub-)set of class labels.
MLC tasks can be found in various domains such as text categorization \cite{schapire2000boostexter,nam2014large,mencia2008efficient}, image processing \cite{cabral2011matrix,xue2011correlative}, video annotation \cite{qi2007correlative}, music classification \cite{sanden2011enhancing}, bioinformatics \cite{barutcuoglu2006hierarchical}, and medicine \cite{heider2013multilabel}.

Generalizing the single-label classification setting, MLC methods often perform a problem transformation, reducing the original MLC task to a (set of) single-label classification problem(s) \cite{tsoumakas2009mining,DBLP:journals/tkde/ZhangZ14}.
Therefore, such problem transformation methods can be seen as a kind of meta-learner that can be configured with one or multiple SLC methods as a base learner.
Moreover, since there are also meta-learners for MLC that require MLC methods in turn as base learners, this results in a nested configuration structure.
The nesting causes the search space over the configuration options for MLC methods to be several orders of magnitude larger than search spaces for SLC or regression, thus, representing a particular challenge for AutoML systems \cite{wevertpami2021}.
Additionally, compared to SLC, solution candidates in MLC are often more expensive to execute. This requires AutoML systems that work in a trial-and-error fashion to be even more efficient.

The motivation for considering the AutoML problem in this thesis is rooted in the broader context of ''On-The-Fly Computing`` \cite{happe2013fly} which is also the name of the collaborative research center (CRC) 901.
It deals with the automatic provision of IT services that are composed and configured on the fly out of base services which in turn are available on worldwide markets.
Focusing on services that provide machine learning functionality, the role of this work is to provide a domain-specific configuration algorithm that is able to configure and tailor machine learning services to the data provided by a user.

In this thesis, we devise a novel approach to AutoML based on hierarchical task network planning \cite{ghallab}, a technique from the field of AI planning, which can naturally model hierarchical dependencies between algorithms as well as hyper-parameters.
Hence, it is flexible enough to represent complex machine learning pipelines and learner structures, as it is necessary for MLC.
Employing a best-first search for a greedy traversal of the resulting search space model, our approach compares well with state-of-the-art AutoML systems in the SLC setting and especially scales with the increased search space complexity in the MLC case.

In Section~\ref{sec:thesis-structure}, we give an overview of the structure and the contributions of this thesis. Section~\ref{sec:running-example} presents a running example.

\section{Thesis Structure}\label{sec:thesis-structure}
The remainder of the thesis is structured as follows.
Chapter~\ref{ch:general-background} is dedicated to the methodological and general background for this thesis introducing fundamental concepts of automated machine learning and multi-label classification.

This Ph.D.~thesis consists of the following contributions, which are presented in Chapters~\ref{ch:mlplan} to \ref{ch:tpami2}, respectively:
\begin{enumerate}[noitemsep]
    \item\fullcite{DBLP:journals/ml/MohrWH18}
    \item\fullcite{wever2018ml}
    \item\fullcite{wever2019automating}
    \item\fullcite{wevertpami2021}
    \item\fullcite{libre}
    \item\fullcite{wever2018ensembles}
    \item\fullcite{mohrtpami2021}
\end{enumerate}
In principle, these contributions can be divided into two parts.

In the first part of this thesis (Chapter~\ref{ch:mlplan} to Chapter~\ref{ch:tpami1}, contributions 1.-4.), we devise a novel approach to AutoML that leverages techniques from AI planning to model the space of machine learning pipelines in terms of a search tree amenable to standard graph search algorithms, such as a best-first search.
In contrast to previous works, this method can naturally reflect hierarchical dependencies of machine learning pipelines, respectively a multi-label classifier, in the search space model, and additionally allows one to systematically search the space via a global search.
After a first version, which is able to configure machine learning pipelines consisting of one learner and at most one preprocessor, which is competitive to the state of the art, we demonstrate the flexibility and scalability of the approach, configuring machine learning pipelines of unlimited length.
Subsequently, we transfer this method to the MLC setting and show that our approach is flexible and efficient enough to scale with the more complex search space.
Moreover, we compare our approach to other state-of-the-art AutoML approaches that we either adapt to the MLC setting or that have already been explicitly proposed for MLC.
In an extensive empirical study, we find that our method is indeed very well suited for the MLC setting and compares favorably with the other considered methods.

The second part of this thesis is concerned with the configuration of learners to increase their effectiveness (Chapters~\ref{ch:libre} and \ref{ch:ndea}, contributions 5 and 6) and how meta-learning can improve the efficiency of AutoML systems (Chapter~\ref{ch:tpami2}, contribution~7).
To this end, we first consider the internal structure of a classifier, so-called \textit{nested dichotomies}, which recursively decomposes the original learning problem into smaller sub-problems, thereby forming a tree-like model structure.
We demonstrate that optimizing this tree structure can lead to improved generalization performance and thus increase the effectiveness of this learner.

Second, we consider \textit{binary relevance learning}, a multi-label classifier that transforms an MLC problem into a set of binary classification problems, one per label, and employs SLC methods for the problems obtained.
While previous literature configures the SLC method jointly for all labels, we show that the effectiveness of this learner can be increased by tailoring the choice of the SLC method to each label.

Third, we show how meta-learning for predicting the runtime of machine learning pipelines can be helpful to render AutoML systems more efficient.
More specifically, we provide empirical evidence for significant savings in wasted computation time for AutoML systems that evaluate candidate pipelines on the given data and impose a limit on the time for this evaluation.
The presented method succeeds in using runtime prediction to avoid timeouts in the evaluation of machine learning pipelines.

In Chapter~\ref{ch:future-direction}, the thesis is concluded, and an outlook on future directions and open questions is given.
Last but not least, since this work has been done in the context of the aforementioned CRC 901 ``On-The-Fly Computing'', Chapter~\ref{ch:otf-computing} is dedicated to the vision of \textit{On-The-Fly Machine Learning}, which refers to the on-demand provision of customized machine learning services.

\section{Running Example}\label{sec:running-example}

\newcommand{\pictureScale}{0.42}
\begin{figure}[t]
     \centering
     \begin{subfigure}[b]{\pictureScale\textwidth}
         \centering
         \includegraphics[width=\textwidth]{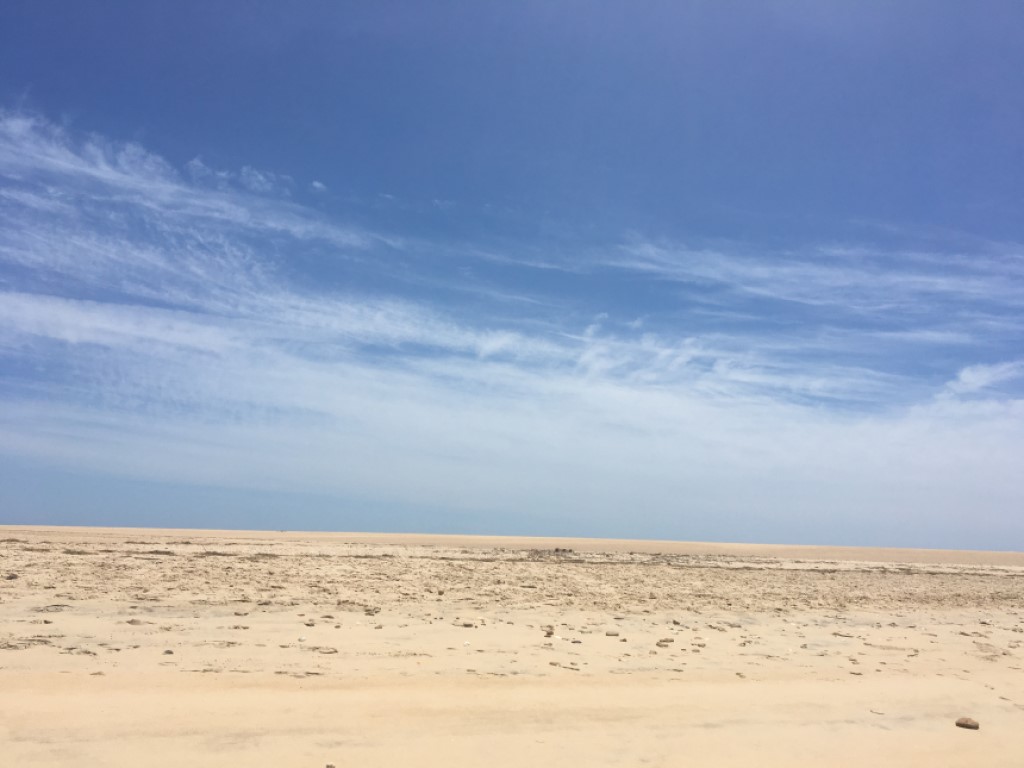}
         \caption{\texttt{BEACH}}
         \label{fig:example-slc-beach}
     \end{subfigure}
     \hfill
     \begin{subfigure}[b]{\pictureScale\textwidth}
         \centering
         \includegraphics[width=\textwidth]{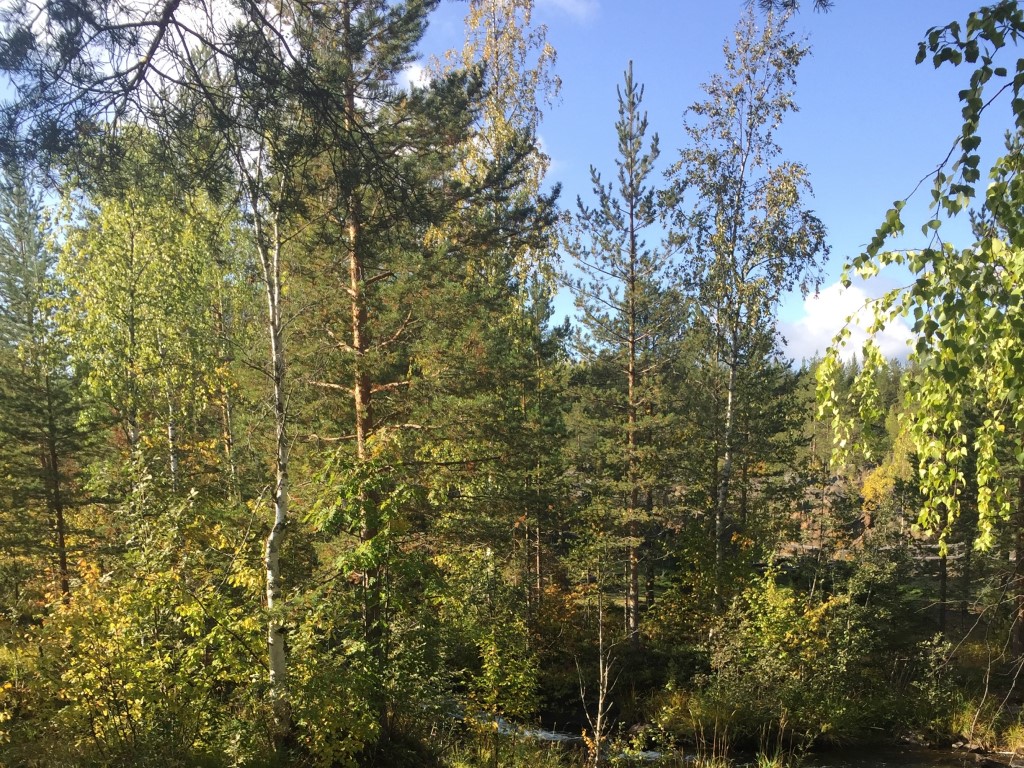}
         \caption{\texttt{FOREST}}
         \label{fig:example-slc-forest}
     \end{subfigure}
     
     \begin{subfigure}[b]{\pictureScale\textwidth}
         \centering
         \includegraphics[width=\textwidth]{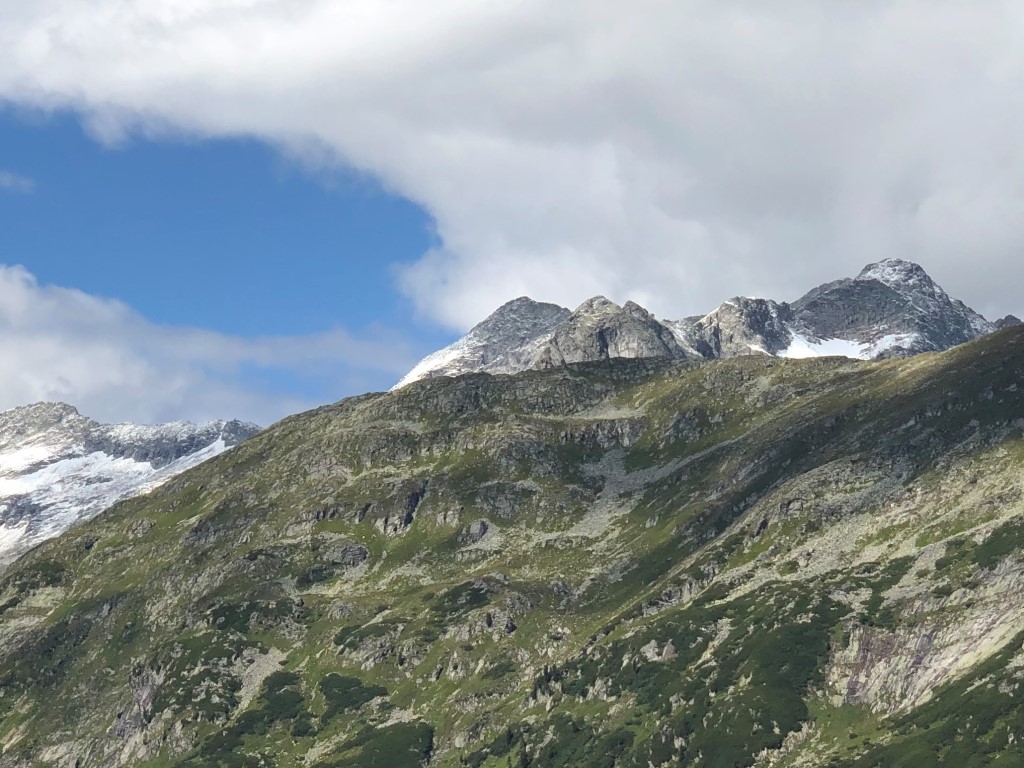}
         \caption{\texttt{MOUNTAIN}}
         \label{fig:example-slc-mountain}
     \end{subfigure}
     \hfill
     \begin{subfigure}[b]{\pictureScale\textwidth}
         \centering
         \includegraphics[width=\textwidth]{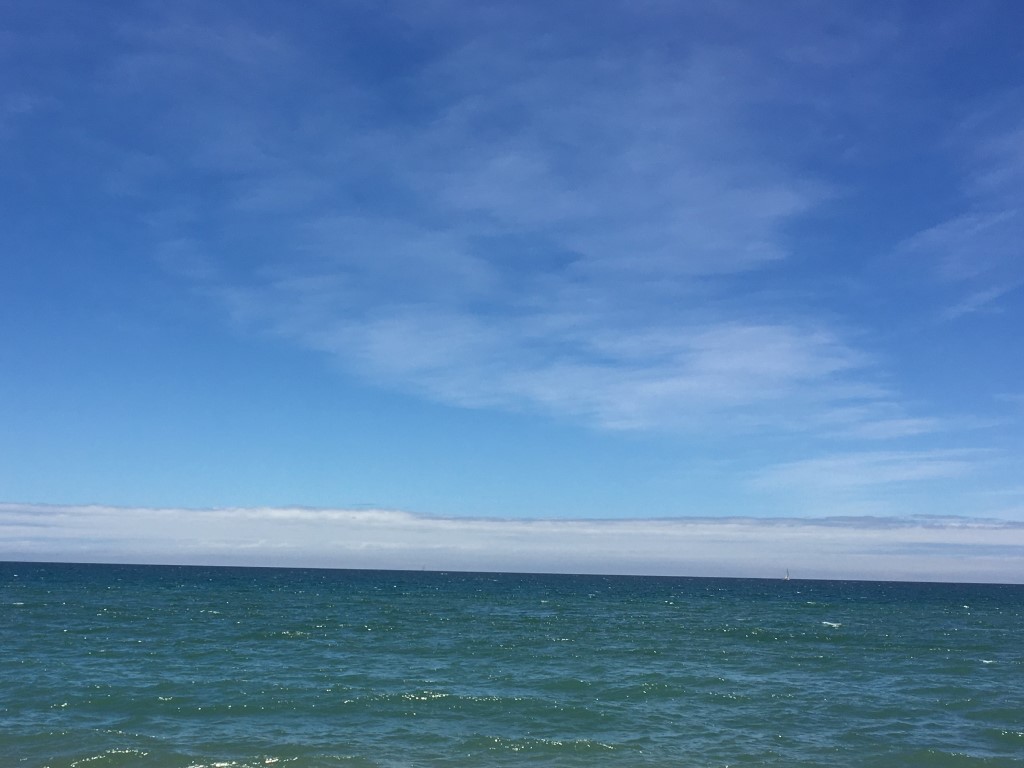}
         \caption{\texttt{SEA}}
         \label{fig:example-slc-sea}
     \end{subfigure}
     
     \begin{subfigure}[b]{\pictureScale\textwidth}
         \centering
         \includegraphics[width=\textwidth]{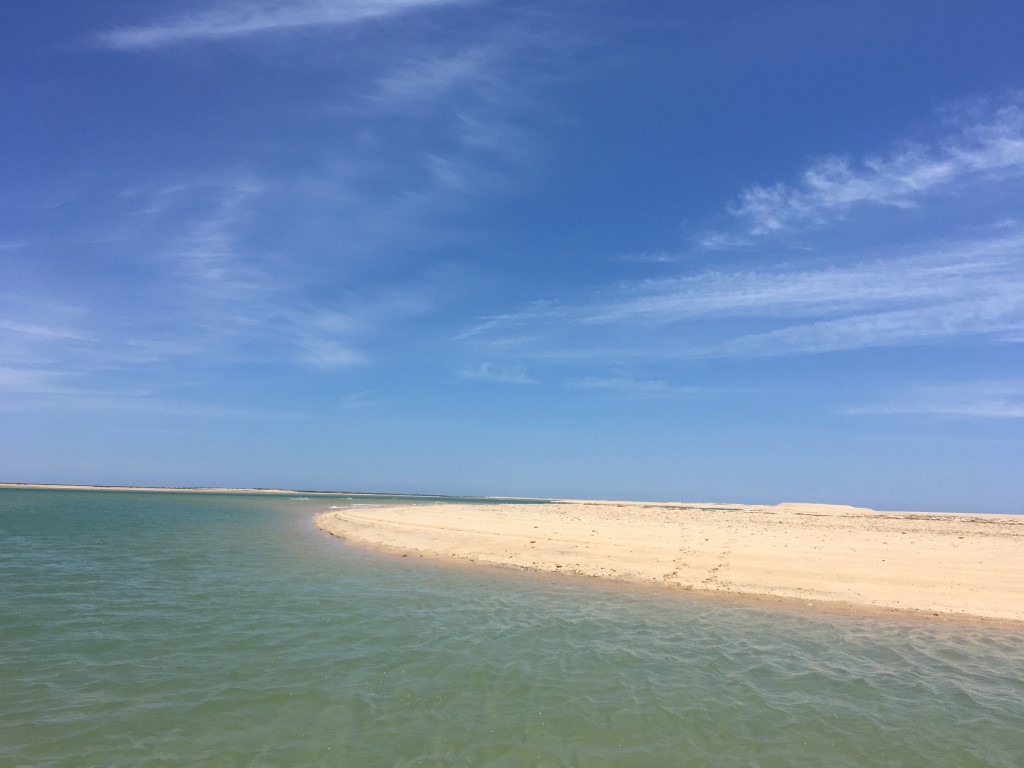}
         \caption{\texttt{BEACH SEA}}
         \label{fig:example-mlc-bs}
     \end{subfigure}
     \hfill
     \begin{subfigure}[b]{\pictureScale\textwidth}
         \centering
         \includegraphics[width=\textwidth]{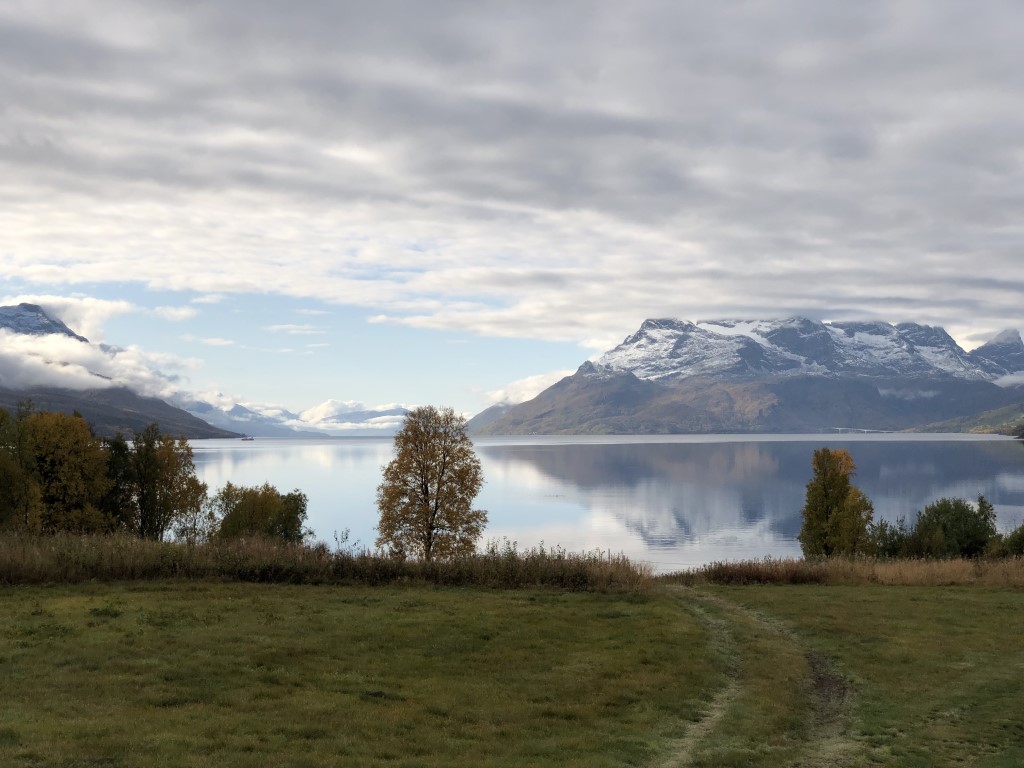}
         \caption{\texttt{MOUNTAIN SEA}}
         \label{fig:example-mlc-ms}
     \end{subfigure}
     
     \begin{subfigure}[b]{\pictureScale\textwidth}
         \centering
         \includegraphics[width=\textwidth]{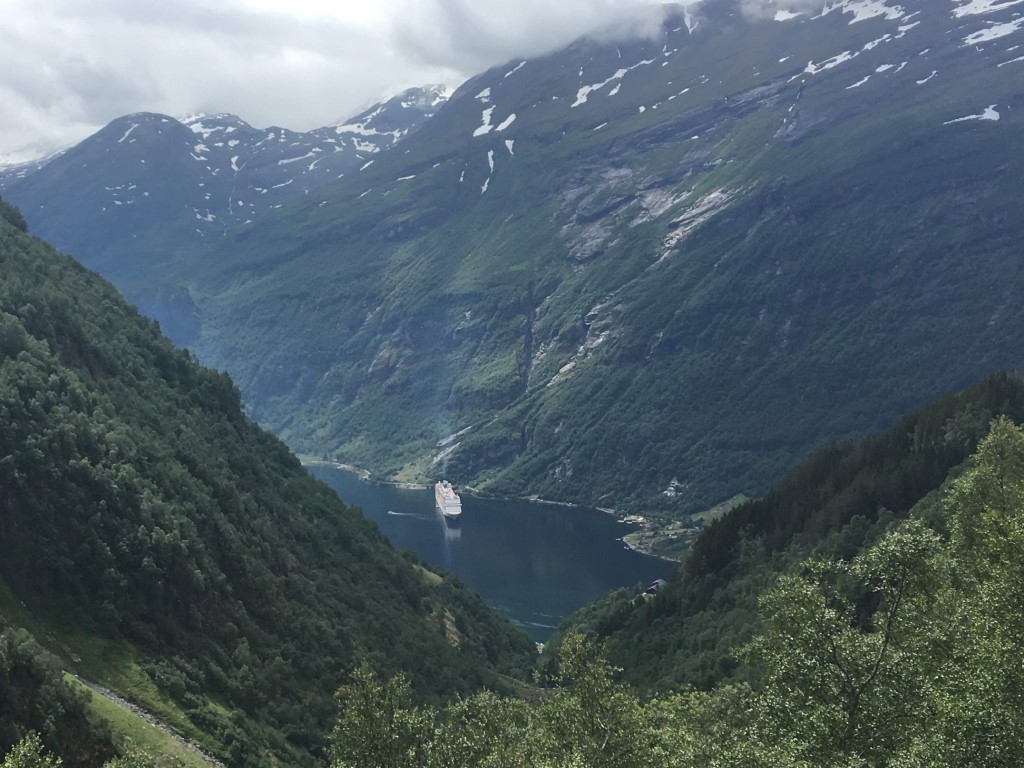}
         \caption{\texttt{FOREST MOUNTAIN SEA}}
         \label{fig:example-mlc-fms}
     \end{subfigure}
     \hfill
     \begin{subfigure}[b]{\pictureScale\textwidth}
         \centering
         \includegraphics[width=\textwidth]{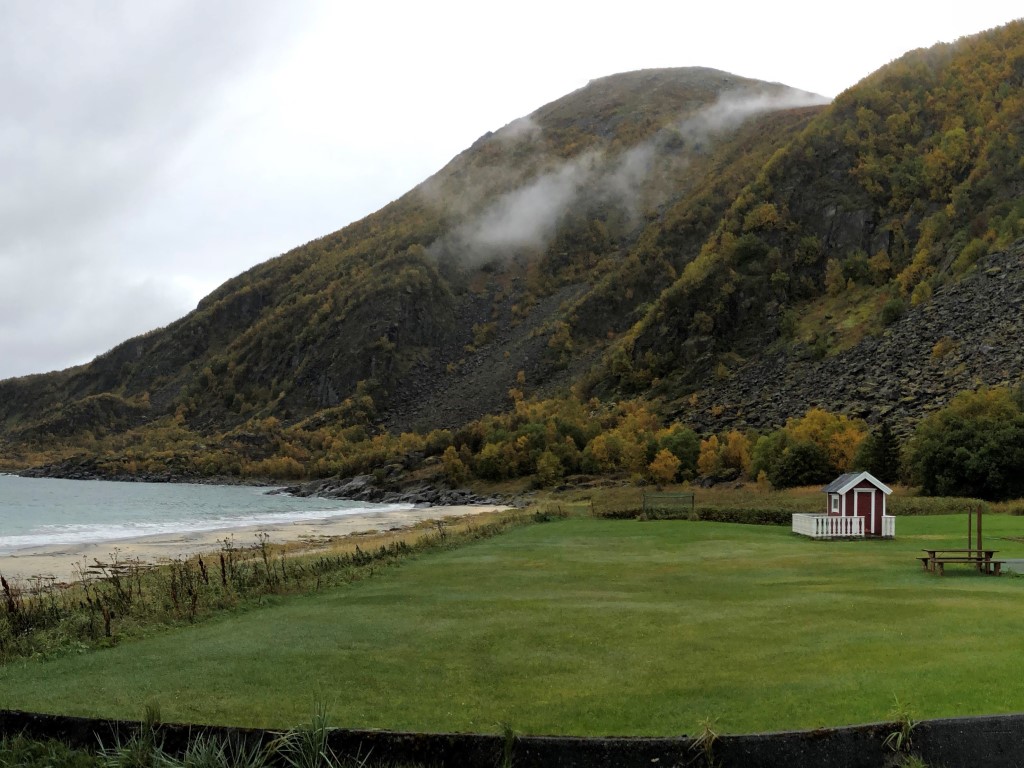}
         \caption{\texttt{BEACH FOREST MOUNTAIN SEA}}
         \label{fig:example-mlc-bfms}
     \end{subfigure}
     
        \caption{Each of the landscape pictures is associated with class labels \texttt{BEACH}, \texttt{FOREST}, \texttt{MOUNTAIN}, and \texttt{SEA}. While the first four pictures can be related to one label exclusively, more than one class label is relevant for the last four pictures. The corresponding sets of labels are detailed in the captions.}
        \label{fig:example}
\end{figure}
Throughout the preliminaries in the following chapter, a continuous example is used to illustrate definitions and concepts.
Suppose we are provided a set of landscape pictures such as shown in Figure~\ref{fig:example}, and we want to classify them by what is shown.
Suppose that we want to label each picture with the help of the class labels $\mathbb{L} = \{\texttt{MOUNTAIN}, \texttt{SEA}, \texttt{FOREST}, \texttt{BEACH} \}$, depending on what is visible on the picture.

When looking at the pictures in Figures~\ref{fig:example-slc-beach} to \ref{fig:example-slc-sea}, each of them can clearly be associated with one of these class labels. More specifically, Figure~\ref{fig:example-slc-beach} shows a \texttt{BEACH}, Figure~\ref{fig:example-slc-forest} a \texttt{FOREST}, Figure~\ref{fig:example-slc-mountain} a \texttt{MOUNTAIN}, and Figure~\ref{fig:example-slc-sea} the \texttt{SEA}.

However, considering pictures as shown in Figures~\ref{fig:example-mlc-bs} to \ref{fig:example-mlc-ms}, it is not that clear anymore.
These images can no longer be assigned to a unique class label since they can be associated with multiple class labels simultaneously.
In Figure~\ref{fig:example-mlc-bs}, both \texttt{BEACH} \textit{and} \texttt{SEA} is shown, whereas \texttt{MOUNTAIN} \textit{and} \texttt{SEA} are visible in Figure~\ref{fig:example-mlc-ms}. Furthermore, Figure~\ref{fig:example-mlc-fms} pictures three of the four class labels, namely \texttt{FOREST}, \texttt{MOUNTAIN}, and \texttt{SEA}. Figure~\ref{fig:example-mlc-bfms} can be associated with the full set of class labels.

\chapter{Preliminaries}\label{ch:general-background}
In this chapter, we provide a general overview of the field of automated machine learning (Section~\ref{sec:automl}) and the learning problem of multi-label classification (Section~\ref{sec:mlc}).

\section{Introduction to Automated Machine Learning}\label{sec:automl}

Automated machine learning (AutoML) refers to the vision of automating the process of selecting and configuring machine learning algorithms, composing them into so-called \textit{machine learning pipelines}, tailored to a given task, i.e., a dataset and a target loss function.
While AutoML was primarily intended to meet the substantial increase in demand for machine learning functionality, it indeed benefits experts in the field as well.
Especially in multi-label classification, the variety of options to configure a multi-label classifier is overwhelming (cf.~Section~\ref{ssec:mlc-configuration}). Hence, a systematic and targeted optimization appears to be hardly possible, even for experts.

The formal definition of the AutoML problem, also referred to as combined algorithm selection and hyper-parameter optimization (CASH) \cite{autoweka}, as the name suggests, is composed of the respective individual optimization problems: algorithm selection (AS) and hyper-parameter optimization (HPO).

Given an instance space $\mathcal{X}$ and a target space $\mathcal{Y}$, let
\begin{itemize}[noitemsep,topsep=-16pt]
\item $\mathcal{A} = \{A^{(1)}, \ldots, A^{(n)}\}$ be a set of algorithms  with corresponding hyper-pa\-ra\-meter spaces $\Lambda^{(1)},\ldots,\Lambda^{(n)}$,
\item $\mathcal{D}= (X,Y) \subset (\mathcal{X}^N \times \mathcal{Y}^N) \subset \mathbb{D}$ a labeled data set from the data set space $\mathbb{D}$,
\item and $\mathcal{L}: \mathcal{Y} \times \mathcal{Y} \mapsto \mathbb{R}$ a target loss function to be minimized.
\end{itemize}
In the context of machine learning, an algorithm $A: \Lambda \times \mathbb{D} \times \mathcal{X}^S \fromto \mathcal{Y}^S$ refers to a learning algorithm that can be configured with a hyper-parameter configuration $\lambda \in \Lambda$ (we write $A_{\lambda}$ in the following).
Furthermore, $A$ takes a data set $\mathcal{D} \in \mathbb{D}$ and a batch of instances $X_\text{test} \subset \mathcal{X}^S$ of size $S$ as arguments.
The data set $\mathcal{D}$ is used by $A$ internally as training data to induce a hypothesis $\widehat{h}$ from some hypothesis space $\mathcal{H} \subseteq \{ h\mid h: \mathcal{X} \fromto \mathcal{Y} \}$.
This hypothesis $\widehat{h}$ is then used to produce predictions on a batch of test instances $X_\text{test} \subset \mathcal{X}^S$ of size $S$ by applying $\widehat{h}$ to each of the $\vec{x} \in X_\text{test}$ individually.
Eventually, $A$ returns the concatenation of all predictions $\widehat{Y} \in \mathcal{Y}^S$.
%
We further assume that every algorithm $A$ has a default parameterization $\lambda_\text{def} \in \Lambda$.
For convenience, if $\lambda$ corresponds to the default parameterization $\lambda_\text{def}$, we write $A$ instead of $A_{\lambda_\text{def}}$.

Then, in AS, the task is to select the most suitable algorithm $A^\ast$ with respect to the target loss function $\mathcal{L}$, i.e.,
\begin{equation}\label{eq:as}
A^\ast \in \underset{A^{(i)} \in \mathcal{A}}{\arg \min}\, \int_{\mathcal{X}\times\mathcal{Y}} \mathcal{L}(\vec{y}, A^{(i)}(\mathcal{D}, \vec{x}))\, P(\vec{x},\vec{y}) \, d\vec{x} d\vec{y} \,\, ,
\end{equation}
where $P(\cdot, \cdot)$ is a joint probability distribution for $\vec{x}$ and $\vec{y}$.
The selection is only made over the finite set of algorithms $\mathcal{A}$.
Thus, in this sense, it does not consider different parameterizations of an algorithm.
However, by considering multiple hyper-parameter configurations of an algorithm $A$ again as a distinct algorithm \cite{tornede2020extreme}, it is possible to include the optimization of hyper-parameters at least to a (very) limited extent.

In turn, the HPO problem fixes an algorithm $A$ in advance and deals with the optimization of the respective hyper-parameters. Given the hyper-parameter space $\Lambda$ of $A$, the optimization problem can be stated as follows.
\begin{equation}\label{eq:hpo}
{\lambda}^\ast \in \underset{{\lambda} \in \Lambda}{\arg \min}\, \int_{\mathcal{X}\times\mathcal{Y}} \mathcal{L}(\vec{y}, A_{\lambda}(\mathcal{D}, \vec{x})) \, P(\vec{x}, \vec{y}) \, d\vec{x} d\vec{y}
\end{equation}
However, the specific algorithm is fixed in this problem, and only its hyper-pa-ra\-meters are considered for optimization.
Therefore, in CASH, as stated first in \cite{autoweka}, the two optimization problems are combined into a joint one, where we seek to find an algorithm $A^\ast$ together with its hyper-parameter configuration $\lambda^\ast$, minimizing the target loss $\mathcal{L}$:
\begin{equation}\label{eq:cash}
A^\ast_{{\lambda}^\ast} \in \underset{A^{(i)} \in \mathcal{A},\, {\lambda} \in \Lambda^{(i)}}{\arg \min} \int_{\mathcal{X}\times\mathcal{Y}} \mathcal{L} \left(  \vec{y}, A^{(i)}_{\lambda} (\mathcal{D}, \vec{x}) \right) \, P(\vec{x}, \vec{y}) \, d\vec{x} d\vec{y} \, .
\end{equation}

In the subsequent Section~\ref{sec:ml-pipelines}, we deal with the concept and different shapes of machine learning pipelines, which are subject to optimization in AutoML.
Furthermore, since the introduction of the CASH problem in 2013 \cite{autoweka}, a variety of methods have been developed more or less following a general schema which is described in Section~\ref{ssec:automl-framework}.
These methods can be divided into three major groups: Reducing the CASH problem to the HPO problem, using grammar-based search approaches, and leveraging meta-learning techniques. We provide details on these three groups in Sections ~\ref{ssec:automl-hpo}, \ref{ssec:automl-grammar}, and \ref{ssec:automl-meta-learning} respectively.
Going beyond CASH, we discuss a sub-field of AutoML called neural architecture search (NAS) in Section~\ref{ssec:automl-nas}, which deals exclusively with the optimization of neural networks.
For a more detailed and in-depth overview of AutoML, we refer the interested reader to surveys shedding light on the field from diverse perspectives \cite{vanschoren2018meta,yao2018taking,zoller2019benchmark,he2021automl}.

\subsection{Machine Learning Pipelines}\label{sec:ml-pipelines}
\begin{figure}
    \centering
    \includegraphics[width=\textwidth]{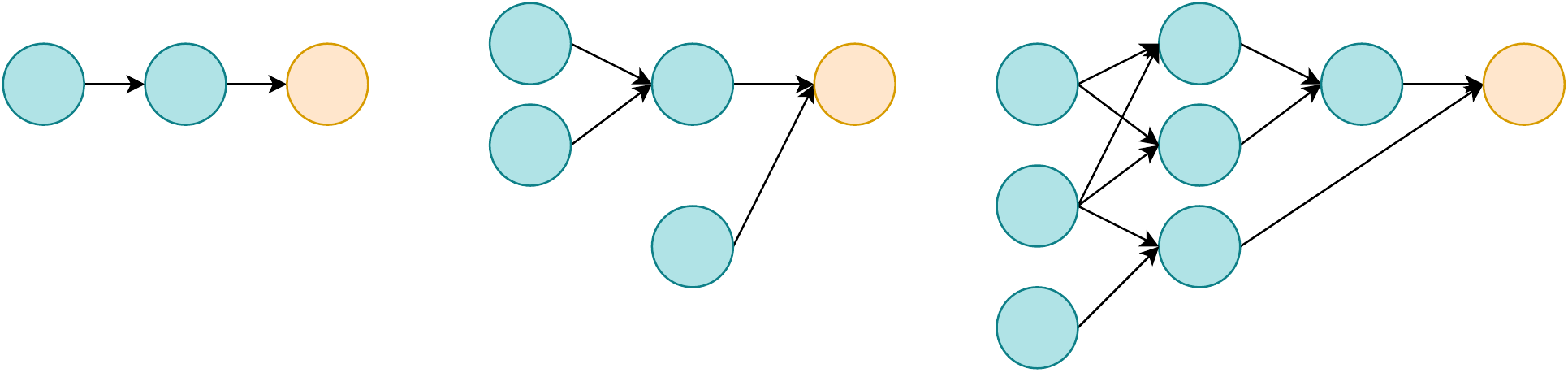}
    \caption{Visualization of different machine learning pipeline topologies. On the left-hand side, a sequential pipeline is shown. The center of the figure presents a tree-shaped pipeline topology, and on the right-hand side, the pipeline structure represents a directed acyclic graph.}
    \label{fig:automl-pipeline-topologies}
\end{figure}

To achieve the best possible performance for a given data set, several machine learning algorithms are often combined into a so-called \emph{machine learning pipeline}. In its simplest form, a machine learning pipeline consists solely of a learning algorithm, and the provided data set is directly used for training. However, depending on the shape of the data and properties of the data and the learning algorithm, some preprocessing of the data might become desirable.

For example, in the case of image tagging, as described in Section~\ref{sec:running-example}, the pictures may need to be preprocessed to make the data amenable to machine learning algorithms since most of these algorithms cannot deal with image data directly. In this case, we would need a machine learning pipeline including an algorithm transforming images into, for instance, a vector representation. Furthermore, it may prove beneficial to ``manipulate'' the images beforehand, e.g., by applying a grey-scale filter, Gaussian filter, or an edge detection algorithm. What kind of preprocessing is needed depends on the data, the task, and also the learning algorithm. For example, neural networks may be directly applied to the pictures, whereas a support vector machine will probably require a more sophisticated preprocessing to perform well.
However, preprocessing is not only required for such unstructured data.
If the learning algorithm cannot deal with nominal features, but the data contains such features, it is inevitable to preprocess the data making it amenable to this learning algorithm. In this case, a preprocessing algorithm could be applied to encode the nominal features in terms of numeric values, e.g., by mapping each attribute value to a specific number.

Modern machine learning libraries, such as WEKA \cite{eibe2016weka}, sckit-learn \cite{pedregosa2011scikit}, or mlR \cite{bischl2016mlr}, comprise a variety of such preprocessing algorithms.
In fact, even learning algorithms can be used as a preprocessor to augment the dataset by predictions of this learning algorithm as yet another feature.
An application of preprocessing algorithms can be done both sequentially and in parallel.

Figure~\ref{fig:automl-pipeline-topologies} shows different types of pipeline structures: A sequential, tree-shaped, and a DAG-shaped\footnote{DAG: directed acyclic graph.} pipeline are visualized.
When executing a pipeline, as a first step, as many copies of the data are created as there are algorithms without a predecessor, i.e., nodes with no incoming edge. After applying the respective algorithm, the preprocessed data is forwarded to the algorithms that are referenced via an outgoing edge. If there are multiple incoming edges, the union of all incoming data is taken. Eventually, the preprocessed data arrives at the learning algorithm, which, in the figure, is highlighted in orange.

\subsection{General Structure of AutoML Systems}\label{ssec:automl-framework}

\begin{figure}
    \centering
    \includegraphics[width=\textwidth]{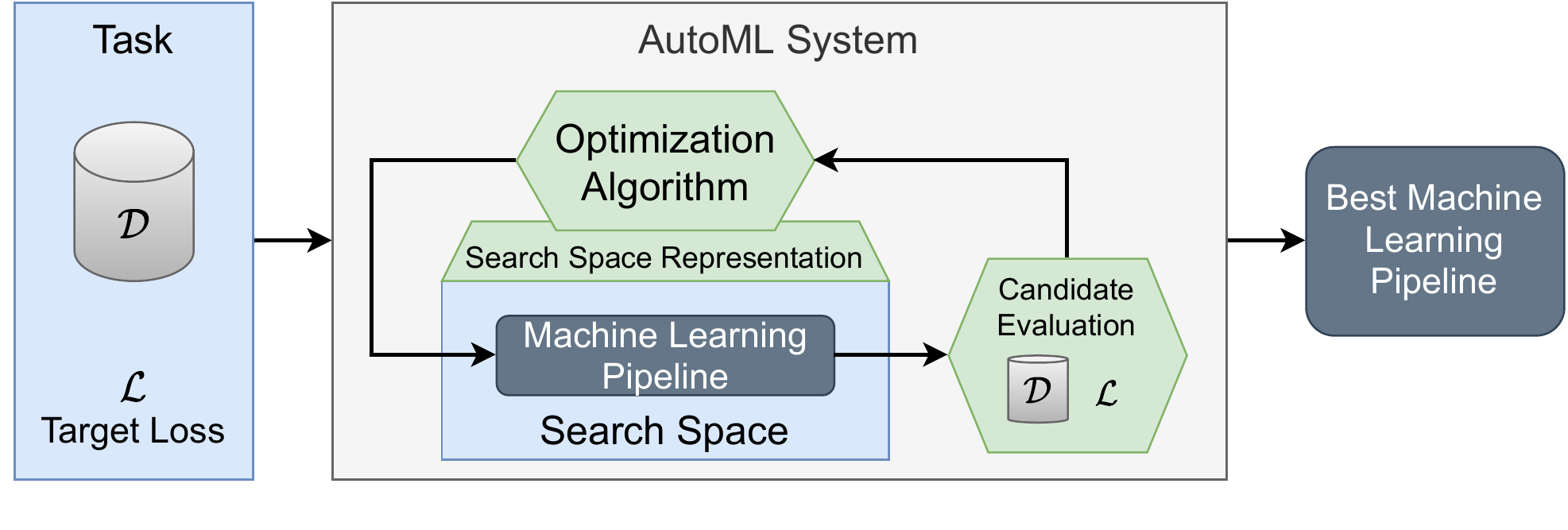}
    \caption{Generic illustration of the AutoML framework. Receiving a task as an input containing a training data set $\mathcal{D}$ and a target loss function $\mathcal{L}$, the AutoML system aims to identify a machine learning pipeline that generalizes well beyond the provided training data. To this end, AutoML systems usually comprise three major components: a search space representation, an optimization algorithm operating on this search space representation, and a candidate evaluation module to assess the solution quality of candidates. Typically, the candidate evaluation uses the provided dataset and the target loss to estimate a candidate's generalization performance.}
    \label{fig:automl-framework}
\end{figure}

The majority of AutoML systems consist of three main components: A specification of the search space that encodes what solution candidates are available, an optimization algorithm, and a function to evaluate solution candidates.
Provided a task as an input, which is specified via a data set $\mathcal{D}$ and a target loss function $\mathcal{L}$, the optimization algorithm traverses the search space to find a machine learning pipeline that is most suitable for the given task.
Eventually, the best-found machine learning pipeline is returned to the user.
An illustration of this general structure is shown in Figure~\ref{fig:automl-framework}.

Generally speaking, the search space representation describes which algorithms are available for selection, which hyper-parameters they expose, and how they can be combined into a machine learning pipeline.
Note that the search space representation is a \textit{model} of the actual search space, i.e., some information or properties of the search space might not be reflected in the search space representation.
Hence, the search space representation can also artificially add information or structures to make the optimization algorithm more effective or efficient, e.g., by grouping similar algorithms.
Typically, the search space representation is designed by an expert and is an integral part of the respective AutoML system.

The optimization algorithm operates on the search space representation and seeks to identify the most suitable solution candidate.
For strategically traversing the search space, the optimization algorithm requires some kind of feedback on the solution quality of candidate machine learning pipelines.
As specified in the problem of Equation~\eqref{eq:cash}, we seek the machine learning pipeline that generalizes best, i.e., minimizes the risk.
However, since the out-of-sample error cannot be computed, we estimate a machine learning pipeline's generalization performance with the help of the provided data set $\mathcal{D}$.
To this end, once commonly splits the data set (randomly) into training $\mathcal{D}_\text{train}$, which is used by the algorithm to induce a hypothesis $\widehat{h}$, and validation data $\mathcal{D}_\text{val} = (X_\text{val}, Y_\text{val})$ for estimating the quality of predictions.
The observed loss on the validation data is returned as feedback to the optimization algorithm.
In fact, we thus solve the optimization problem
\begin{equation}\label{eq:cash-val}
A^\ast_{\lambda^\ast} \in \underset{A^{(i)} \in \mathcal{A},\, \lambda \in \Lambda^{(i)}}{\arg \min} \mathbb{E}_{\mathcal{D}_\text{train}, \mathcal{D}_\text{val}} \left[ \mathcal{L} \left(  Y_\text{val}, A^{(i)}_{\lambda} (\mathcal{D}_\text{train}, X_\text{val}) \right) \right]
\end{equation}
as a surrogate for the actual problem \eqref{eq:cash}. Here, the expectation accounts for any randomness of algorithms $A_{\lambda}^{(i)}$, the data itself, and the precise way how the algorithms are validated, e.g., via k-fold cross-validation, a hold-out set, etc.

As already mentioned, to determine the loss of algorithm $A_{\lambda}^{(i)}$, it is typically executed on the given data.
Specifically, this means that executable implementations of the respective algorithms are required to run the optimization process.
Therefore, the search space descriptions are usually based on software libraries for machine learning such as scikit-learn \cite{pedregosa2011scikit,varoquaux2015scikit}, WEKA \cite{eibe2016weka}, or in the case of multi-label classification MEKA \cite{MEKA} and Mulan \cite{tsoumakas2009mining}.

\subsection{Reduction to Hyper-Parameter Optimization}\label{ssec:automl-hpo}

While the very first attempt to AutoML was made in \cite{DBLP:conf/nips/BergstraBBK11} employing a random search, the first AutoML systems going beyond a simple random search perform a reduction from CASH to HPO \cite{autoweka,autoweka2,autosklearn}, making it amenable to well researched and sophisticated HPO approaches, such as Bayesian optimization, multi-armed bandits, or genetic algorithms.
To this end, the choice of algorithms is encoded as yet another (categorical) hyper-parameter and combined into a joint hyper-parameter vector together with all the hyper-parameters exposed by the respective algorithms.

In terms of the HPO problem \eqref{eq:hpo}, we consider an (artificial) meta-algorithm $\widehat{A}$ with hyper-parameter space $\widehat{\Lambda} = \{\lambda_A \mid \lambda_A \in \{1, \ldots n\} \} \times \Lambda^{(1)} \times \ldots \times \Lambda^{(n)}$.
Depending on the concrete choice of the algorithm via the categorical hyper-parameter $\lambda_A$, only a small subset of hyper-parameters is relevant for optimization, namely the ones belonging to the selected algorithm $A^{(\lambda_A)}$.
In the following, we refer to these hyper-parameters interchangeably as \textit{active} hyper-parameters.
Which hyper-parameters are considered active is usually modeled in an auxiliary conditional structure.

\textbf{Random search} Arguably, the simplest and most straightforward way of approaching the HPO problem is via random search \cite{DBLP:conf/nips/BergstraBBK11,gil2018p4ml, ledell2020h2o}.
In random search, solution candidates $\lambda \in \widehat{\Lambda}$ are sampled randomly and evaluated with respect to the given task.
Despite its simplicity, a random search can yield reasonably good results \cite{gijsbers2019open}.
Another advantage of random search is that it can be parallelized arbitrarily, making it particularly appealing for distributed systems and cloud environments.

\textbf{Bayesian Optimization} Bayesian optimization (BO) \cite{DBLP:journals/corr/abs-1807-02811} is an often-used approach for HPO, not least because of its efficiency and theoretical guarantees of convergence.
Generally speaking, BO is a technique for optimizing black-box functions $f: U \fromto \mathbb{R}$, where $U$ denotes some input domain (e.g., $\mathbb{R}^N$) and the evaluation of $f$ is assumed to be expensive.
Furthermore, it does not require additional knowledge about $f$, such as gradients, convexity, linearity, etc.

\begin{figure}[t]
\center
\includegraphics[width=.5\textwidth]{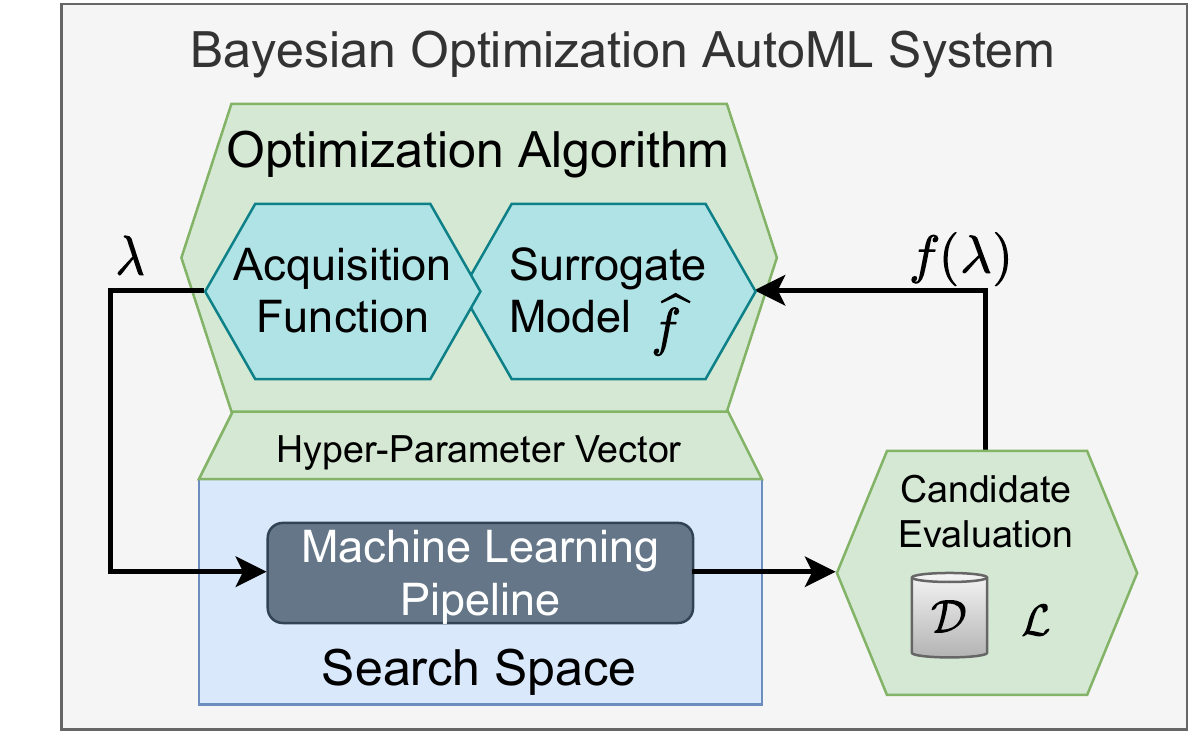}%
\caption{Illustration of an AutoML system employing Bayesian optimization. Solution candidates are represented in terms of a hyper-parameter vector. The surrogate model $\widehat{f}$ models the actual evaluation function $f$ to be optimized. Furthermore, $\widehat{f}$ is used by the acquisition function to decide, which hyper-parameter configuration $\lambda$ to sample next. Prior to evaluation, the chosen $\lambda$ is translated into a machine learning pipeline. Then, $f(\lambda)$ augments the set of observations of $f$, which in turn updates the surrogate model $\widehat{f}$.}%
\label{fig:bo}%
\end{figure}

The optimization procedure of BO, as depicted in Figure~\ref{fig:bo}, comprises mainly two components: (i) a surrogate $\widehat{f}$ of $f$, which is cheap to evaluate, and (ii) an \textit{acquisition function} forming the basis for decisions on which input to evaluate the actual function $f$ next.
The surrogate is represented by a probabilistic model estimated from the actual function evaluations of $f$ that have been observed so far.
The most commonly used models are Gaussian processes \cite{snoek2012practical}, random forests \cite{smac}, and Tree-structured Parzen Estimator \cite{bergstra2013hyperopt,smac}.
An essential capability of these surrogate models is to provide information about the expected values -- and thus possible optima of $f$ -- and quantify their uncertainty about these predictions. Based on this information, the acquisition function is tasked to trade-off exploration and exploitation to sample $f$ efficiently. Here, exploration refers to ``exploring'' configurations for which the predictions of the surrogate $\widehat{f}$ has high uncertainty, whereas exploitation means to evaluate configurations for which the surrogate model predicts a high performance.

In the context of HPO, the black-box function $f$ of interest is the (estimated) performance of a hyper-parameter configuration in terms of a loss function $\mathcal{L}$.
Producing this information involves fitting and validating the algorithm with the respective hyper-parameter configuration, which can indeed be considered expensive as it can take up to minutes or even several hours. Consequently, the number of function evaluations that can be afforded is generally very limited.

An example of such an acquisition function is the expected improvement (EI) \cite{DBLP:conf/ifip/Mockus77,DBLP:journals/jgo/JonesSW98}.
The basic idea of EI is to evaluate the hyper-parameter configuration next, which maximizes the improvement over the previous best solution candidate, as would be expected according to the surrogate model. Formally, the EI for a hyper-parameter configuration $\lambda$ with respect to the best hyper-parameter configuration $\widehat{\lambda}^\ast$ observed so far can be computed via
\begin{equation}
    \mathit{EI}(\lambda) = \mathbb{E}\left[ \max\left(f(\widehat{\lambda}^\ast) - f(\lambda), 0\right) \right] \, ,
\end{equation} 
where $f(\hat{\lambda}^*)$ is the result of executing $f$ for the hyper-parameter configuration $\hat{\lambda}^*$, i.e., the best outcome observed so far, $f(\lambda)$ is a random variable with an unknown outcome at the time of the computation of $\mathit{EI}(\lambda)$. Practically, in order to compute $EI(\lambda)$, we make use of the surrogate model $\widehat{f}$ to estimate the mean and the variance of $f(\lambda)$.

Based on BO, several approaches to AutoML have been proposed.
Probably the better-known representatives of these AutoML systems are Auto-WEKA \cite{autoweka,autoweka2} and auto-sklearn \cite{autosklearn}.
However, various extensions and enhancements have been proposed \cite{zhang2016flash,feurer2018towards,maher2019smartml,komer2019hyperopt} but also specializations focusing on single algorithms \cite{thomas2018automatic,jin2019auto}.

\textbf{Multi-Armed Bandits}
Formalizing the problem as a multi-armed bandit (MAB) problem \cite{slivkins2019introduction} is another way of tackling the CASH problem as a reduction to HPO.
MAB denotes a sequential decision-making problem in which an agent must repeatedly select options from a given set of alternatives in an online setting.
The metaphor of the eponymous slot machines in casinos associates the available options with the ``arms'' of the slot machine that are ``pulled'', i.e., the respective option is selected.
By pulling an arm, the agent is provided a \textit{reward} signal as feedback on the quality of his or her choice.
Typically, only one arm is pulled at a time, and the goal is to optimize a certain evaluation criterion, e.g., maximizing the cumulative reward.
To achieve such a goal, the agent must carefully balance exploration, to possibly find a better arm, but at the same time risking to draw arms yielding less reward, with exploitation, to take advantage of the arms already known to yield high rewards.
 
Instantiating the MAB problem for HPO, each arm represents a hyper-parameter configuration, and the aim of the agent is to identify the best arm.
Moreover, the reward signal for pulling an arm is provided by evaluating the corresponding hyper-parameter configuration for a particular budget $b$, e.g., time or number of observations used for training.
The agent may choose and adapt the budget $b$ for a hyper-parameter configuration over time. Consequently, this requires that the candidate evaluation function $f$ needs to be parameterizable in the budget $b$.

\begin{figure}
    \centering
    \includegraphics[width=.5\textwidth]{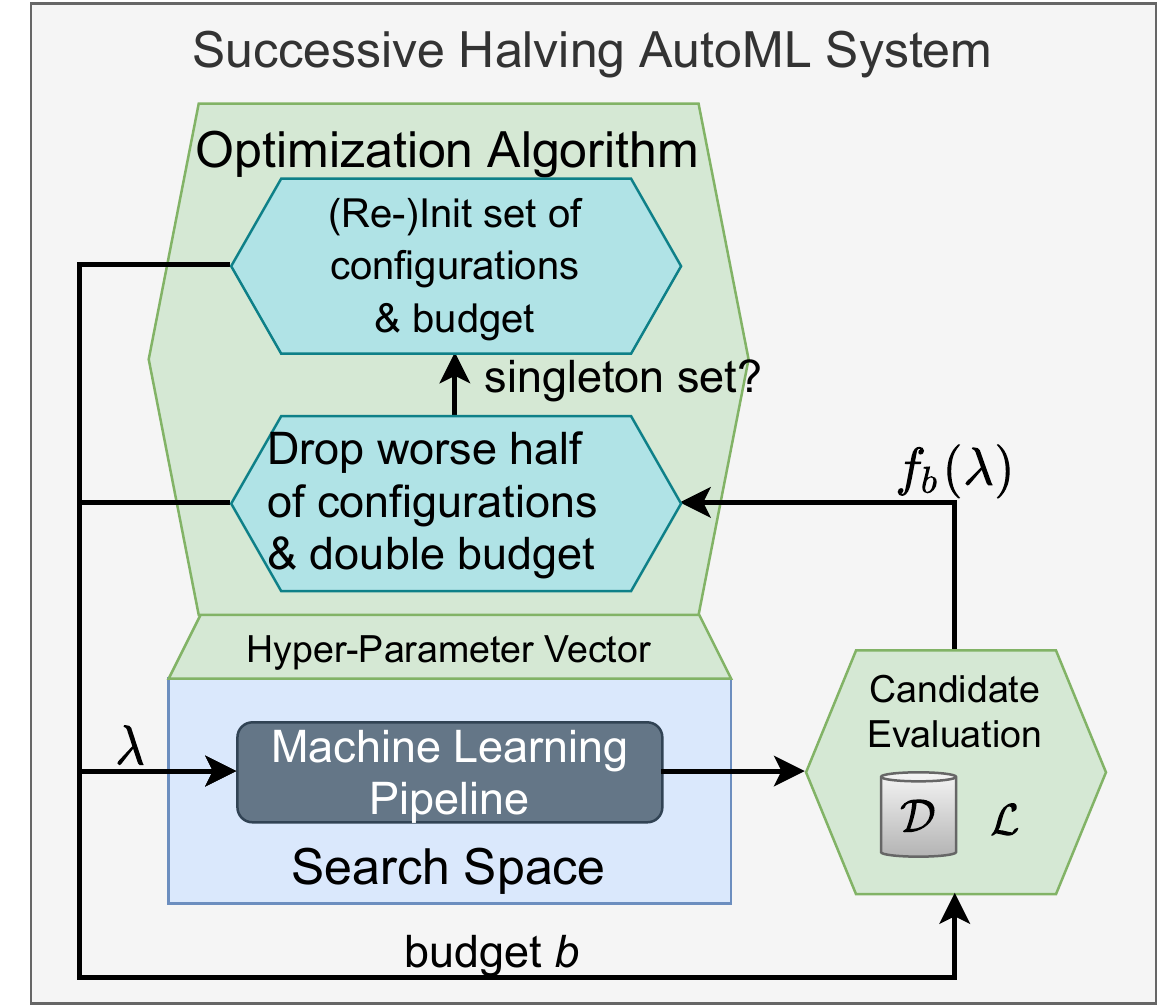}
    \caption{Illustration of an AutoML system, employing a successive halving algorithm for optimization with a budgeted candidate evaluation function $f_b(\cdot)$. After picking an initial set of configurations and initial budgets, the algorithm iteratively evaluates configurations for the current budget $b$, discards the worse half of configurations, and re-evaluates the remaining for an increased budget. This process is run multiple times, varying the initial budget and the size of the initial set of configurations.}
    \label{fig:automl-sh}
\end{figure}

Based on the MAB setting, an approach to HPO is proposed in \cite{DBLP:conf/aistats/JamiesonT16} applying the successive halving (SH) algorithm \cite{DBLP:conf/icml/KarninKS13}.
Theoretically and empirically, the proposed approach is shown to yield good performance.
Initially sampling a representative set of hyper-parameter configurations at random, as the name suggests, it successively disregards half of the worse performing configurations, based on the observed average performance after a certain amount of the available budget is reached.
An illustration of this approach is provided in Figure~\ref{fig:automl-sh}.

Note that the set of all hyper-parameter configurations is usually huge, if not infinite, which is already the case as soon as real-valued parameters are considered.
However, in \cite{DBLP:conf/aistats/JamiesonT16}, this problem is addressed by sampling a predefined number of configurations, thus, only presenting a finite set of configurations to SH.
After each halving, the remaining configurations are evaluated for an increased budget.
In this way, SH allows for wasting less budget on poorly performing configurations and for allocating more budget for the more promising ones, finding the best arm after $\ceil{\log_2(N)}$ iterations, where $N$ is the number of initially sampled hyper-parameter configurations.

However, defining the value of $N$ has a practical impact on the choice of the final arm \cite{DBLP:journals/jmlr/LiJDRT17}.
Mainly, this is due to (i) sampling too few configurations may lead to missing good ones, and (ii) sampling too many configurations decreases the budget initially assigned for evaluation such that configurations that perform better on larger budgets are disregarded early.
Hence, the number of initially sampled configurations is a crucial hyper-parameter to the optimization algorithm.
Depending on the maximum allocatable budget, in \cite{DBLP:journals/jmlr/LiJDRT17}, the authors propose a heuristic, which they dub Hyperband, to run SH for different numbers of initially sampled configurations repeatedly. As an optimization algorithm of an AutoML system, Hyperband performs competitively to systems based on Bayesian optimization \cite{autoband}.

\begin{figure}
    \centering
    \includegraphics[width=.6\textwidth]{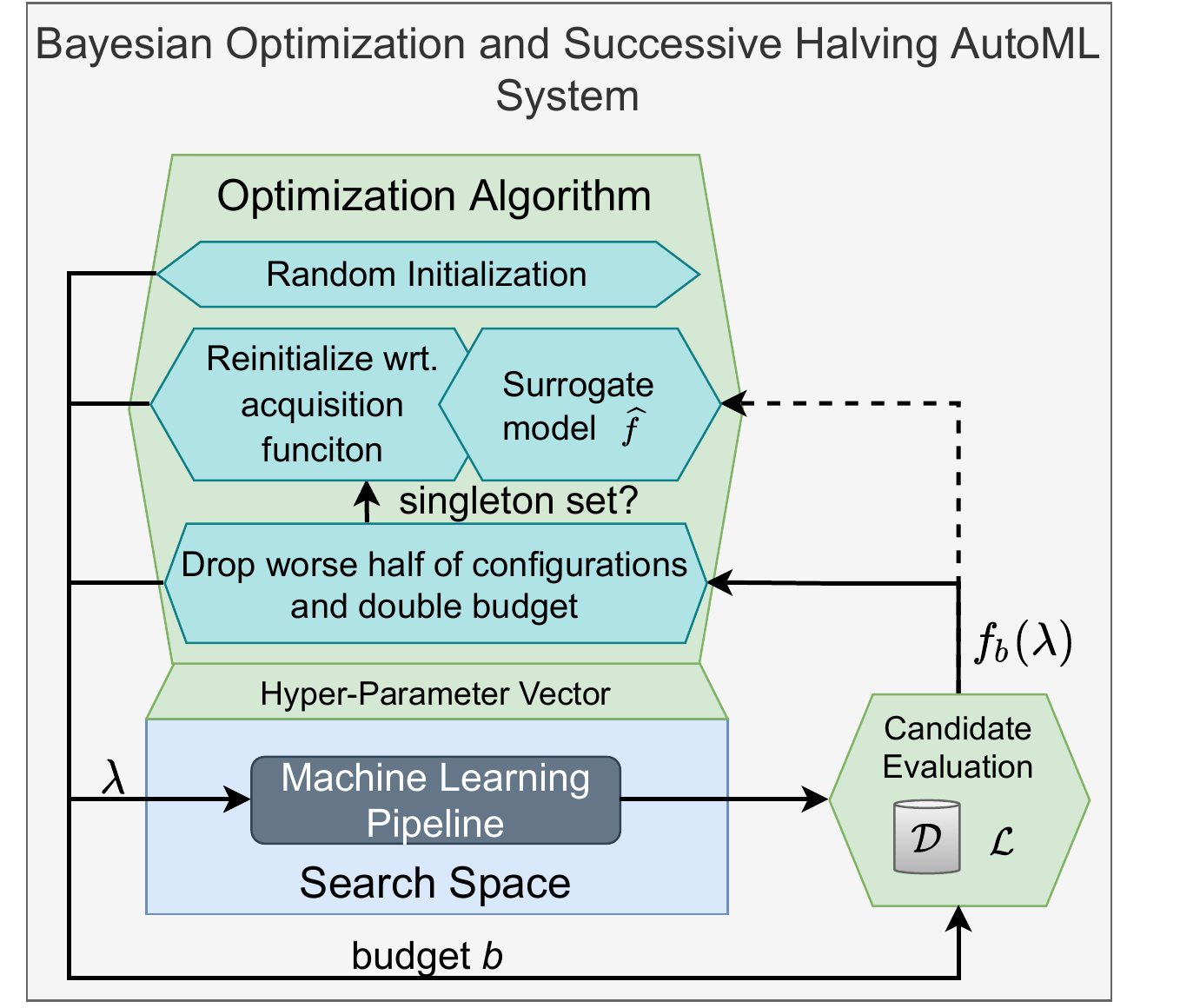}
    \caption{Illustration of an AutoML system combining Bayesian optimization and Hyperband (BOHB) into a hybrid optimization algorithm leveraging the best out of the two worlds: Successive halving as in Hyperband together with a model-based sampling of promising candidates when reinitializing the set of configurations.}
    \label{fig:bohb}
\end{figure}

\textbf{Hybrid Bayesian Optimization and Multi-Armed Bandits}
A significant disadvantage of Hyperband is that knowledge about already observed performances from previous iterations is not incorporated for sampling new sets of initial configurations.
To overcome this limitation, the idea of Hyperband is integrated with the Bayesian optimization approach leading to a hybrid algorithm called Bayesian Optimization and Hyperband (BOHB) \cite{bohb}.
In fact, the random sampling procedure of Hyperband is (mostly) substituted by sampling configurations according to the surrogate $\widehat{f}$ and an acquisition function.
Due to technical reasons, some of the configurations are still sampled at random to ensure convergence. The combination of BO and Hyperband is visualized in Figure~\ref{fig:bohb}.
It has been applied to address the AutoML problem in \cite{swearingen2017atm} and \cite{feurer2018practical}.

\textbf{Others}
Beyond the discussed methods, other approaches reducing the AutoML problem to HPO exist, for example, by using standard genetic algorithms \cite{de2017towards,wever2018ensembles}, or via grid search to successively build a stacking ensemble \cite{erickson2020autogluon}.

Another interesting approach is presented in \cite{liu2020admm}, leveraging the alternating direction method of multipliers (ADMM) algorithm, where a bipartition of the numeric and categorical hyper-parameters is created. Thereby, the original optimization problem is decomposed into simpler sub-problems (concerning the number of variables), containing only variables of a single type. Furthermore, the ADMM framework allows for introducing additional (external) constraints on the optimization problem, such as fairness or robustness constraints.

\subsection{Grammar-Based Search}\label{ssec:automl-grammar}

The reduction of the CASH to the HPO problem has a considerable disadvantage with regard to more flexible machine learning pipelines.
When working with a hyper-parameter vector as a search space representation, its size is fixed, and therefore also a maximum length or number of components within the pipeline is predetermined.
Another branch of AutoML systems emerged to overcome this limitation that uses a grammar-based representation of the search space.
This representation does not impose any limitations on the maximum length or the number of components contained in a machine learning pipeline.
Furthermore, it allows for describing different shapes of machine learning pipelines -- e.g., sequential, tree-shaped, and even in the shape of a directed acyclic graph -- compared to a fixed structure as predefined in the reduction to HPO.

Another advantage of grammar-based approaches is that they can also naturally represent recursive dependencies.
The configuration of learning algorithms, especially meta-learning algorithms, can become quite complex.
This is because it requires the configuration of a base algorithm which in turn is a learning algorithm exposing hyper-parameters and needs to be configured by the AutoML system as well.

\textbf{Grammar-Based Genetic Programming}
Grammar-based optimization approaches are pretty prevalent in the field of evolutionary algorithms, especially in the sub-field of grammar-based genetic programming (GGP) \cite{koza1995survey,ggp}.
In the original sense, genetic programming is about synthesizing software with the help of evolutionary algorithms \cite{cramer1985representation}, assessing the fitness $f(\cdot)$ of individuals, for example, in terms of test coverage to verify that the program works as expected.
Instead of representing individuals in terms of gene strings, GGPs employ a formal language or grammar to describe the space of solutions, and individuals are derived based on this grammar.
More precisely, individuals in GGPs describe a derivation tree for this grammar.

\begin{figure}
    \centering
    \includegraphics[width=.5\textwidth]{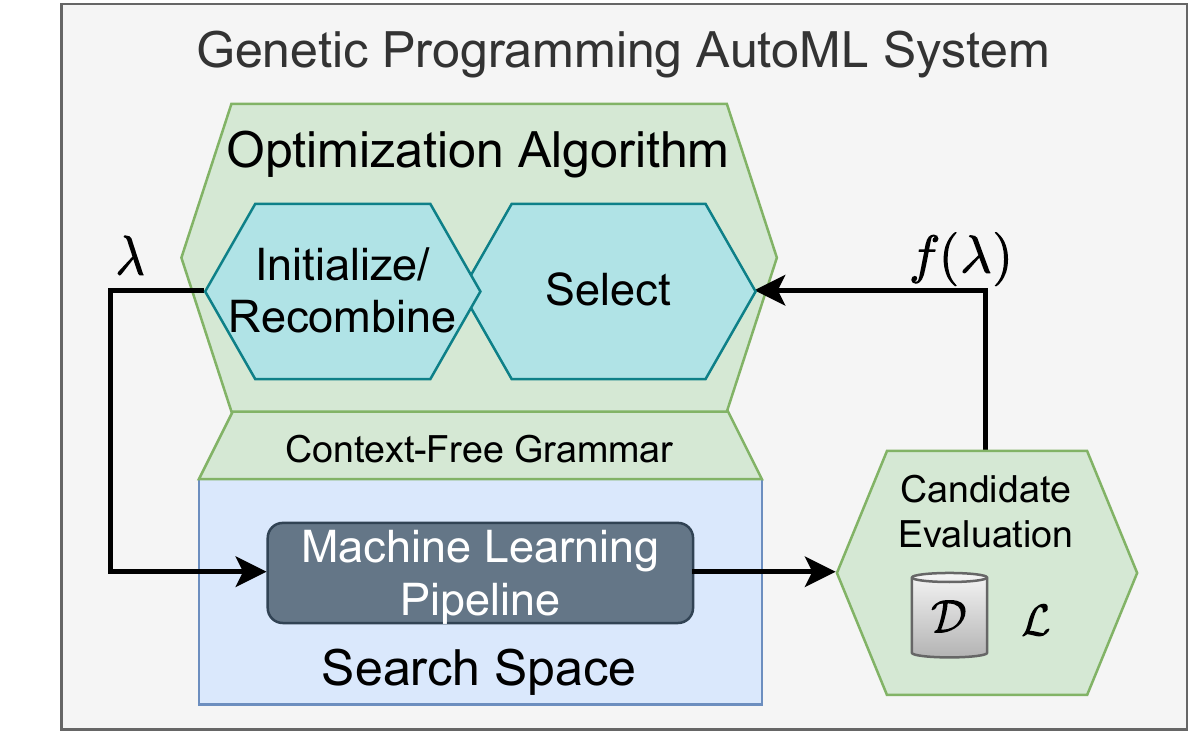}
    \caption{Illustration of an AutoML system employing genetic programming as an optimization algorithm. As common in evolutionary algorithms, genetic programming maintains a population of solution candidates, also referred to as individuals. The fitness values $f(\cdot)$ computed for individuals are used to select more promising ones and use them as input for recombination operators. The distinctive feature of genetic programming is that individuals are represented in the form of trees. In the AutoML domain, these trees are derivation trees of some context-free grammar describing the space of potential machine learning pipelines.}
    \label{fig:automl-ggp}
\end{figure}

Using GGPs for AutoML has the advantage that the grammar allows much more flexible structures for machine learning pipelines than a fixed-length hyper-parameter vector as needed when applying BO. An illustration of a genetic programming AutoML system is given in Figure~\ref{fig:automl-ggp}.
AutoML systems based on genetic programming support tree-shaped pipelines \cite{tpot,recipe} and even support representations for more complex structures such as pipelines involving feature extraction \cite{tornede2021coevolution} or pipelines exhibiting a directed acyclic graph structure \cite{chen2018autostacker}. Moreover, GGPs also natively allow for recursive structures as in the case of configuring multi-label classifiers \cite{de2018automated}.
However, as is typical for evolutionary algorithms, GGPs maintain a population of individuals of a specific size.
On the one hand, parallel processing of the individuals is trivial, but on the other hand, the size of the population is also a crucial hyper-parameter that needs to be chosen with care. 
While a smaller population allows for faster processing of generations, initializing a population that covers the search space in a sufficiently representative way is difficult.
A larger population, in turn, may provide better coverage of the search space, but on the other hand, it also increases the computational costs for evaluating the offspring of each generation.
Moreover, the next generation does not begin until the previous one is complete, so single expensive-to-evaluate individuals may stall the entire evolution. 
This is especially an issue when dealing with larger data sets where the evaluation of single candidates does not take seconds but rather minutes or hours.

To overcome these limitations and to improve the scalability of TPOT \cite{tpot}, it has been extended to leverage successive halving for fitness evaluations \cite{parmentier2019tpot} and to include a feature set selector as another algorithm that can be used in a candidate pipeline \cite{le2020scaling}.
Another work addresses the issues of waiting for a generation to finish before beginning a new one employing an asynchronous GGP variant \cite{gijsbers2019gama}.
More specifically, this GGP continuously recombines and evaluates individuals without the synchronization barriers of generations.

\textbf{AI Planning and Graph Search}
Hierarchical task network (HTN) planning \cite{ghallab} originates from the field of automated planning and scheduling, sometimes simply referred to as AI planning, and deals with the automated production of \textit{plans}, i.e., sequences of actions, that are typically executed by an (intelligent) agent. In HTN planning, the idea is to structure the search space hierarchically based on a logic language and operators defined on that language.

A plan or a solution is derived in HTN planning by refining so-called tasks arranged in a partially ordered set of tasks, also referred to as task network. Tasks can either be \textit{complex} or \textit{primitive}. While the latter represent actionable items, i.e., actions that the agent can execute, the former needs to be refined until only primitive tasks are left. Complex tasks can be viewed as a composition of simpler tasks and thus need to be refined via so-called \textit{methods} into those simpler tasks, which can be again complex or primitive.
The way the search space is described is strongly reminiscent of context-free grammars, with complex tasks corresponding to non-terminals, primitive tasks to terminals, and methods to production rules.
However, the difference here is that methods can impose additional constraints in the form of preconditions that must be satisfied for the method to be applicable to a particular state.

\begin{figure*}[t]
\center
\includegraphics[width=\textwidth]{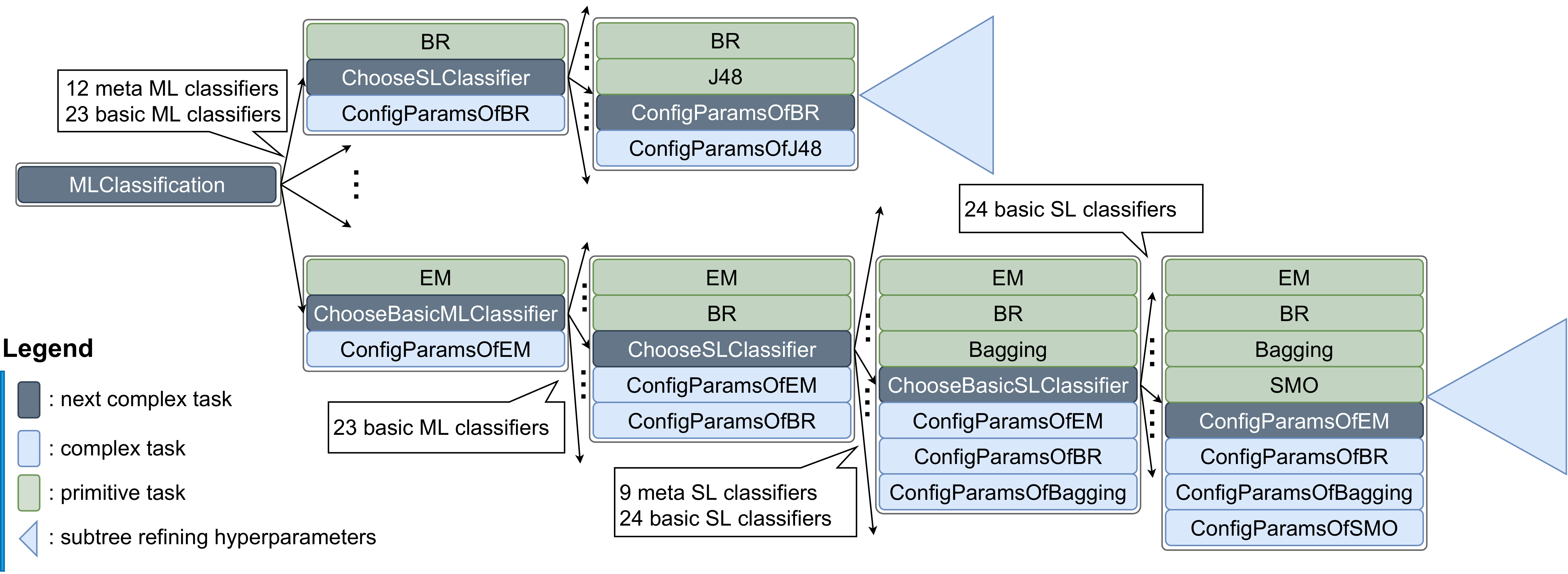}%
\caption{Creation of pipelines with hierarchical planning. \textbf{Top:} A binary relevance (BR) learning classifier is configured with a decision tree, which is called J48 in WEKA, as a base learner. \textbf{Bottom:} First, a meta multi-label classifier expectation maximization (EM) is selected and configured with a binary relevance learning classifier as a base learner, which in turn employs a bagged SMO classifier ensemble as a base learner for the individual labels.}%
\label{fig:derivations}%
\end{figure*}

To solve HTN-planning problems, they are usually reduced to graph search problems, making them accessible to standard graph search algorithms, such as breadth-first or depth-first search.
This reduction is often made by \textit{forward-decomposition} \cite{ghallab}, which selects the \textit{first} complex task in the network of a node to be refined.
The node's successors are then obtained, considering each method that can be applied to refine the selected task.
In this way, every inner node of the induced search graph corresponds to a plan prefix, i.e., the primitive tasks that have already been fixed, and to a task network containing the remaining complex tasks that still need to be refined.
Hence, the start node represents an empty plan prefix and a task network containing the initial complex task.
Leaf nodes, in turn, represent complete plans and empty rest networks.
In Figure~\ref{fig:derivations} a concrete example is depicted, where the initial complex task ``MLClassification'' is decomposed in plan prefixes.
Already fixed primitive tasks of the plan prefixes are shown in green, whereas complex tasks are colored in blue.
The next complex task to be refined is highlighted in dark blue.

In the context of automating data mining and machine learning, HTN planning, respectively an extension of HTN planning called programmatic task network (PTN) planning \cite{ptnplanning}, has been used for modeling algorithm choices and hyper-parameter values in terms of primitive tasks and introducing complex tasks as abstraction layers for grouping different kinds of algorithms and defining structures such as the topology of a machine learning pipeline \cite{DBLP:conf/ecai/KietzSBF12,wever2017automatic,DBLP:journals/ml/MohrWH18,wever2018ml,katz2020exploring}. In \cite{DBLP:journals/ml/MohrWH18,wever2018ml} (Chapters~\ref{ch:mlplan} and \ref{ch:automl1}), we propose an AutoML system named ML-Plan, applying a best-first search to the resulting search graph which requires each node to be assigned a heuristic score.
Since an inner node of the search graph corresponds to an incomplete specification of a machine learning pipeline and thus, cannot be evaluated as a candidate, random completions are drawn to leaf nodes.
The machine learning pipelines represented by the respective leaf nodes are then evaluated, and the observed performances are aggregated at the inner node, e.g., taking the minimum. 
The number of random completions affects the trade-off between exploration and exploitation.
While a larger number promotes exploration, a smaller number reinforces the greediness of the search and consequently exploitation.

Beyond single-label classification, HTN planning was also used in extensions of ML-Plan to other types of tasks, such as remaining useful lifetime estimation in the problem domain of predictive maintenance \cite{tornede2020pdm}, and multi-label classification (Chapters~\ref{ch:automl2} and \ref{ch:tpami1}) \cite{wever2018automated,wever2019automating,wevertpami2021}.

Representing the search space as a search graph or a search tree, other heuristic graph search methods can be successfully applied as optimization algorithms of AutoML systems. For example, in \cite{rakotoarison2019automated} a Monte-Carlo tree search is applied along with a progressive widening strategy \cite{chaslot2008progressive} to make the high branching factor manageable.
In \cite{plmcts}, the authors suggest the use of a Plackett-Luce model \cite{ceberio2013plackett} to improve Monte-Carlo tree search for single-player games, e.g., AutoML.
An adaptation of AlphaZero \cite{silver2017mastering}, which also employs a Monte-Carlo tree search, is proposed in \cite{drori2018alphad3m}.
Another approach combines reinforcement learning for selecting algorithms with Bayesian optimization for configuring the hyper-parameters of machine learning pipelines \cite{sun2019reinbo,lin2019ml}.
During a hyper-parameter optimization phase, the best performance is fed back as a reward to the reinforcement learning algorithm.

\subsection{Meta-Learning}\label{ssec:automl-meta-learning}
A major criticism of AutoML systems is that they are too resource-intensive in terms of both computational power as well as time. Following the common design scheme, as shown in Figure~\ref{fig:automl-framework}, AutoML systems heavily rely on actually executing candidate machine learning pipelines in order to estimate their generalization performance for the given task. Besides the high computational costs, the execution of machine learning pipelines takes some time ranging from milliseconds to hours or even days -- depending on the data and algorithms contained in the candidate pipelines. Hence, also the AutoML process can be pretty time-consuming.

\begin{figure}
    \centering
    \includegraphics[width=.5\textwidth]{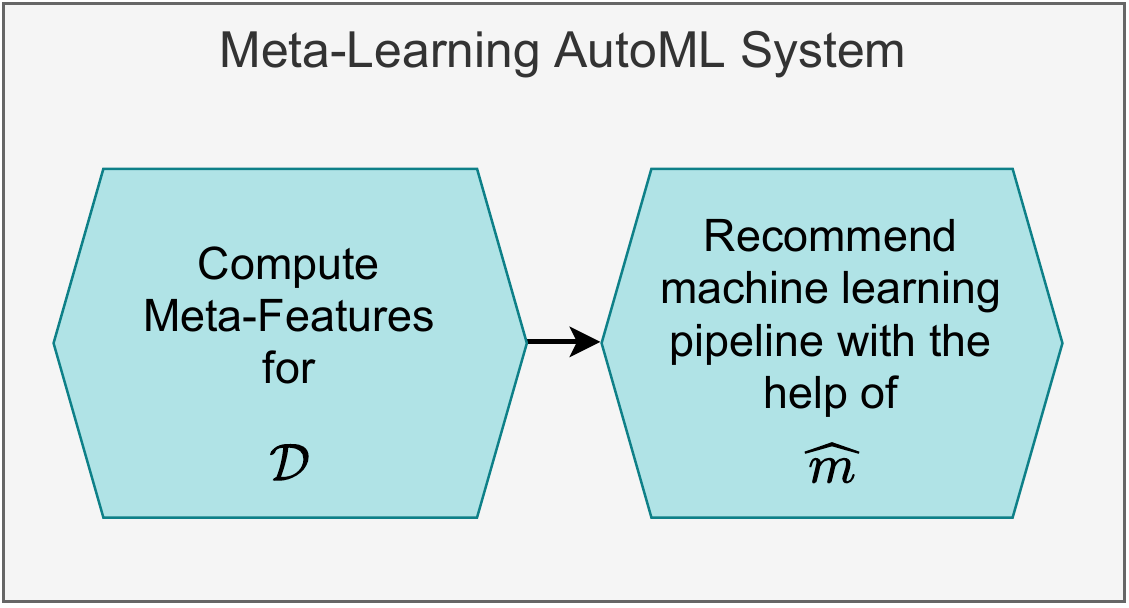}
    \caption{AutoML systems using meta-learning for recommending machine learning pipelines usually require a feature representation of datasets which needs to be computed for the given dataset $\mathcal{D}$, before the meta-model $\widehat{m}$ can be queried to obtain a recommendation. Note that this AutoML system does not involve a candidate evaluation, which would require the machine learning pipelines to be executed for the given data set $\mathcal{D}$.}
    \label{fig:meta-learning}
\end{figure}

One possible way to address these issues is through meta-learning.
Inspired by the way humans learn, meta-learning (also known as ``learning to learn'') aims to accumulate experience (metadata) on already solved tasks to accomplish future tasks faster and/or more accurately.
Transferred to the AutoML problem, this translates to observing performance data of machine learning pipelines on training data sets and, based on this, estimating a model to make recommendations for unseen data sets.
On the one hand, meta-learning can be used to support and speed up existing AutoML systems, e.g., by first evaluating the best machine learning pipelines known for similar (already examined) datasets, as it is done in \cite{autosklearn,maher2019smartml}.
Moreover, in Chapter~\ref{ch:tpami2}, we advocate the use of meta-learning for predicting runtimes of machine learning pipelines to decide whether a pipeline should be evaluated or not. Avoiding evaluations of machine learning pipelines that take too long significantly improves the efficiency of AutoML systems \cite{mohrtpami2021}.
Alternatively, one may think of substituting the entire AutoML system with a predictor that recommends a machine learning pipeline for a new (unseen) data set.

Rather than treating each dataset independently, as suggested by the original problem statement in Equation~\eqref{eq:cash}, with the help of a trial-and-error-based AutoML system, we seek to learn a mapping $\widehat{\varphi}$ from the space of datasets to the space of algorithms and their parameterizations, as defined in the introduction of Section~\ref{sec:automl}.
$$
    \widehat{\varphi}: \mathcal{D} \fromto \{ (A^{(i)}, \lambda) \mid A^{(i)} \in \mathcal{A}\,\, \text{and} \,\, \lambda \in \Lambda^{(i)} \}
$$
$\widehat{\varphi}$ represents a surrogate of the underlying ground truth function $\varphi$, assigning each dataset the best machine learning pipeline.
Note that, in general, $\varphi$ cannot be observed as the ground truth assignment of a machine learning pipeline to a data set can usually not be determined with absolute certainty since only \textit{estimates} on the generalization performance are observable.

By simplifying the CASH problem to algorithm selection and covering different parametrizations in terms of (many) ``different'' algorithms, the problem can be solved, for example, by probabilistic matrix factorization \cite{fusi2018probabilistic,yang2019oboe} or by leveraging learning algorithms, in turn, to predict performances of machine learning pipelines \cite{tornede2020extreme}.
In the latter work, we use a dyadic feature representation, comprising a feature representation of data sets and a feature representation of algorithms, allowing the model to generalize across both dimensions data sets and algorithm parametrizations. 
While in the first place, this dyadic feature representation improves data efficiency of the meta-learner, i.e., only a few training observations are required to make highly accurate predictions, in principle, it technically allows for making predictions on unknown parametrizations of the already known algorithms.
However, whether these meta-models generalize well to unknown parametrizations is still an open question.

A key role for good generalization performance, and thus, the successful application of such meta-models lies in the informativeness of the so-called meta-features, i.e., features describing properties of the data sets and -- in the case of a dyadic feature representation -- algorithms.
Often, statistical features (such as the number of instances, features, classes, etc.) and performances of fast-to-evaluate learners, also known as landmarking features, are used to describe datasets. A more sophisticated way is proposed in \cite{drori2019automl} leveraging deep language models to transform natural language descriptions of data sets into a numeric feature representation. Although natural language embeddings seem promising \cite{jomaa2021dataset2vec}, coming up with a suitable feature representation is still an open research area.

\subsection{Neural Architecture Search}\label{ssec:automl-nas}

Neural architecture search (NAS) denotes a particular branch of AutoML committed to and specialized in optimizing neural networks, especially on optimizing the topology of neural networks.
Initial work in this sub-field of AutoML uses evolutionary algorithms for optimization and introduces an abstraction layer by defining building blocks also known as \textit{cells} that can be composed to obtain the final architecture \cite{real2017large}.
Although they are evolved from scratch, neural networks optimized via NAS manage to match the performance of highly sophisticated neural networks designed by human experts.
Since the seminal work \cite{real2017large}, a plethora of other approaches and improvements have been proposed to make the search for neural architectures more resource-efficient or to reduce the number of internal parameters while maintaining a competitive generalization performance \cite{wistuba2018deep,zhu2019eena,xie2020weight}. Besides, NAS has also been used for optimizing neural networks for multi-label classification ~\cite{cascadeML}.
For a more comprehensive overview of NAS, we refer the interested reader to overview papers on NAS \cite{elsken2019neural,wistuba2019survey,ren2020comprehensive}.

Although the increased interest in NAS -- presumably due to the success of neural networks in image classification -- seems rather recent, the optimization of neural network architectures has already been a topic of interest in the evolutionary computation community since the 80s and 90s \cite{DBLP:conf/nips/HarpSG89,DBLP:journals/adb/Gruau94}.
For example, in the field of evolutionary robotics, controllers for robots are implemented via neural networks, which are optimized by evolutionary algorithms \cite{nolfi2016evolutionary}.
The optimization involves the internal parameters and the neural network architecture, i.e., individual neurons and connections between neurons.
However, for image classification tasks, the neural network architectures are larger by several orders of magnitude, and a key component for making NAS approaches succeed is to work with the abstract entity of cells instead of the neurons and connecting edges directly.

\section{Introduction to Multi-Label Classification}\label{sec:mlc}
Multi-label classification (MLC) refers to a special form of a multi-target prediction problem \cite{waegeman}, where data points (instances) are associated with binary targets denoting the ``relevance'' or the ``irrelevance'' of a specific property of interest, which is represented by a class label.
Moreover, it generalizes the more common classification tasks of binary and multi-class classification, where only a single such class label is associated with each instance.
In MLC we aim to learn a predictor, mapping instances to \textit{subsets of} (presumably) \textit{relevant labels}. As an example, consider the image tagging example of Section~\ref{sec:running-example} where a picture can be associated with multiple labels, e.g., \texttt{BEACH}, \texttt{FOREST}, \texttt{MOUNTAIN}, and \texttt{SEA}, at the same time.
For a more comprehensive and in-depth overview of multi-label classification, we refer the interested reader to the survey articles \cite{DBLP:reference/dmkdh/TsoumakasKV10} and \cite{DBLP:journals/tkde/ZhangZ14}.

Subsequently, we give a formal definition of the MLC learning problem in Section~\ref{ssec:mlc-problem-setting}.
Furthermore, we elaborate on how single-label classification, i.e., binary and multi-class classification, can be framed as a special instance of MLC in Section~\ref{ssec:mlc-slc}.
In Section~\ref{ssec:mlc-loss-functions}, we discuss various loss functions that are typically used for assessing the quality of predictions. Then, we give an overview of different ways of approaching the MLC learning problem in Section~\ref{ssec:mlc-classifiers}. After elaborating on the concept of label dependence in more detail in Section~\ref{ssec:mlc-label-dependence}, we conclude this introduction by elaborating on the challenge of configuring multi-label classifiers in Section~\ref{ssec:mlc-configuration}.

\subsection{Problem Definition}\label{ssec:mlc-problem-setting}

Let $\mathcal{X}$ denote the space of instances and $\mathbb{L} = \{ l_1, \ldots l_m \}$ a finite set of $m$ labels.
In MLC, we assume each instance $\vec{x} \in \mathcal{X}$ to be (non-deterministically) associated with a subset of labels $L \subseteq \mathbb{L}$ via a joint probability distribution $P(\cdot, \cdot)$ for $\vec{x}$ and $L$.
We call $L$ the set of \textit{relevant labels} and its complement $\mathbb{L} \setminus L$ the set of \textit{irrelevant labels}.

The set of relevant labels $L$ can be conveniently represented in terms of a binary vector $\vec{y} = (y_1, \ldots, y_m$), where $y_i = 1$ if the label $l_i$ is considered to be relevant, i.e., $l_i \in L$.
Alternatively, if a label $l_i$ belongs to the set of irrelevant labels, i.e.,  $l_i \in \mathbb{L} \setminus L$, we set $y_i = 0$.
In terms of this representation, we can denote the set of all possible label combinations by $\mathcal{Y} \defeq \{0,1\}^m$.

Based on this, we can define a multi-label classifier $\vec{h}$ as a function $\vec{h}: \mathcal{X} \fromto \mathcal{Y}$.
This function takes an instance $\vec{x} \in \mathcal{X}$ as input and returns a binary vector (\textit{prediction})
\[
    \vec{h}(\vec{x}) = \big( h_1(\vec{x}), \ldots, h_m(\vec{x}) \big) \, ,
\]
indicating the relevance of each label, where $h_i(\vec{x})$ represents the $i$-th entry of the vector returned by applying $\vec{h}(\cdot)$ to $\vec{x}$. Furthermore, we denote as $$\mathcal{H} \subseteq \{ \vec{h} \mid \vec{h} : \mathcal{X} \fromto \mathcal{Y} \}$$ the hypothesis space specifying the available multi-label classifiers $\vec{h}$.

Provided a finite set of $N$ observations
$$
\mathcal{D}_\text{train} \defeq (X_\text{train}, Y_\text{train}) = \{ (\vec{x}_i, \vec{y}_i) \}_{i=1}^{N} \subset \mathcal{X}^N \times \mathcal{Y}^N
$$
as training data, we aim for inducing such a multi-label classifier $\vec{h}: \mathcal{X} \fromto \mathcal{Y}$, generalizing well beyond this finite set of observations.
More specifically, we seek to find a classifier $\vec{h}^\ast\in \mathcal{H}$ from a hypothesis space $\mathcal{H}$ that, after fitting the training data $\mathcal{D}_\text{train}$, minimizes the risk with respect to a target loss function $\mathcal{L}: \mathcal{Y} \times \mathcal{Y} \fromto \mathbb{R}$ (cf.~Section~\ref{ssec:mlc-loss-functions})
\begin{equation}\label{eq:mlc-problem}
\vec{h}^\ast \in \underset{\vec{h} \in \mathcal{H}}{\arg \min}
\int_{\mathcal{X}\times\mathcal{Y}} \mathcal{L}(\vec{y},\vec{h}(\vec{x})) P(\vec{x}, \vec{y})\, d\vec{x}d\vec{y} \,\, ,
\end{equation}
where $P(\cdot\, ,\cdot)$ refers to a(n) (unknown) joint probability distribution for $\vec{x}$ and $\vec{y}$.

\subsection{Single-Label Classification}\label{ssec:mlc-slc}
Standard classification tasks, in the following referred to as \textit{single-label classification} (SLC), such as binary or multi-class classification problems, can be understood as special cases of MLC.
While limiting the number of labels $m$ to $1$ yields a binary classification task, we can derive the multi-class classification problem by imposing a constraint of mutual exclusiveness on the labels. More specifically, for any set of relevant labels $L \subseteq \mathbb{L}$, we restrict the size of $L$ to be $|L|=1$. Following the notation of the binary vector $\vec{y}$, this constraint can be formulated by the sum of its entries $y_i$ being equal to 1, i.e., $\sum_{i=1}^m y_i = 1$.

\subsection{Loss Functions}\label{ssec:mlc-loss-functions}
Over time, a wide array of loss functions has been proposed to quantify the quality of predictions, many of which generalize or adapt losses that are well-known in the literature on single-label classification (SLC).
This section presents a subset of these loss functions, most commonly used in the literature.
For a more comprehensive overview, we refer the reader to \cite{wu2017unified}.

To estimate the generalization performance of a multi-label classifier, let
\begin{equation}
\mathcal{D}_\text{test} \defeq (X_\text{test}, Y_\text{test}) = \{ (\vec{x}_i, \vec{y}_i) \}_{i=1}^S \subset \mathcal{X}^S\times\mathcal{Y}^S
\end{equation}
be a test set of size $S$.

For convenience, we interpret $Y_\text{test}$ in the following as a matrix
\begin{equation*}
Y =
\begin{pmatrix}
\vec{y}_1\\
\vec{y}_2\\
\ldots\\
\vec{y}_S
\end{pmatrix}
=
\begin{pmatrix}
y_{1,1}  & \ldots & y_{1,m}\\
y_{2,1}  & \ldots & y_{2,m}\\
\ldots & \ldots & \ldots \\
y_{S,1}  & \ldots & y_{S,m}\\
\end{pmatrix}
\end{equation*}
where $y_{i,j}$ denotes the ground truth relevance of a label $j$ for observation $i$.

Similarly, we can write the predictions of a multi-label classifier $\vec{h}$ on $X_\text{test}$ as a matrix $\widehat{Y}$:
\begin{equation*}
\widehat{Y}=
\begin{pmatrix}
\vec{h}(\vec{x}_1)\\
\vec{h}(\vec{x}_2)\\
\ldots\\
\vec{h}(\vec{x}_S)
\end{pmatrix}
=
\begin{pmatrix}
h_1(\vec{x}_1) & \ldots & h_m(\vec{x}_1)\\
h_1(\vec{x}_2) & \ldots & h_m(\vec{x}_2)\\
\ldots & \ldots  & \ldots\\
h_1(\vec{x}_S)  & \ldots & h_m(\vec{x}_S)\\
\end{pmatrix}
=
\begin{pmatrix}
\hat{y}_{1,1} & \ldots & \hat{y}_{1,m}\\
\hat{y}_{2,1}  & \ldots & \hat{y}_{2,m}\\
\ldots & \ldots & \ldots\\
\hat{y}_{S,1}  & \ldots & \hat{y}_{S,m}\\
\end{pmatrix}
=
\begin{pmatrix}
\hat{\vec{y}}_{1}\\
\hat{\vec{y}}_{2}\\
\ldots\\
\hat{\vec{y}}_{S}\\
\end{pmatrix}
\end{equation*}
Here, an entry $\hat{y}_{i,j}$ represents the prediction of $\vec{h}$ for label $j$ of observation $i$. Hence, each row represents the predictions for an observation, and vice versa, each column for a specific label.
Then, an MLC loss function can be defined as $\mathcal{L} : \mathcal{Y}^S \times \mathcal{Y}^S \fromto [0,1]$.\footnote{More generally, MLC loss functions may take a matrix $\mathcal{Y}^S$ for the ground truth labels and a matrix $\mathbb{R}^{S\times m}$ representing the predictions of $\vec{h}$ as arguments since various multi-label classifiers produce label relevance scores ranging in $[0,1]$ instead of sharp assignments of $0$ or $1$. Again other classifiers may yield unbounded relevance scores. Via a threshold $\tau$ such scores can be transformed into 1 if the score exceeds the threshold $\tau$ and 0 otherwise. When dealing with classifiers producing scores, for the sake of simplicity, we assume these scores to be thresholded in the following.}

The generalization or adaptation of SLC losses to the MLC setting can be classified into three categories.
\begin{description}
    \item[macro instance-wise] In macro instance-wise multi-label loss functions, the loss function is first computed for each observation (row) individually and subsequently aggregated across all observations, e.g., by taking the mean. By computing the loss for each observation first, more emphasis is put on predicting all the labels of an instance correctly.
    \item[macro label-wise] Here, the SLC loss function is first computed for each label, i.e., individually for each column. The losses obtained in this manner are then aggregated, e.g., via the mean. Thus, in contrast to the macro instance-wise losses that emphasize making correct predictions for instances, in macro label-wise loss functions, the focus is on predicting labels correctly.
    \item[micro] Lastly, rather than reinforcing better predictions for either instances or labels, any prediction $\hat{y}_{i,j}$ can be considered equally important. To this end, micro loss functions arrange the entries of the matrices $Y$ respectively $\widehat{Y}$ in terms of vectors before computing the loss.
\end{description}

For example, the \textit{subset 0/1} loss is a macro instance-wise generalization of the well-known error rate.
It considers the predicted and expected label subsets as a whole and considers the entire prediction for an instance to be wrong whenever the two sets do not coincide:
\begin{equation}
\mathcal{L}_{0/1} (Y_\text{test}, \widehat{Y}) \defeq \dfrac{1}{S} \sum_{i=1}^S  \big\llbracket \vec{y}_{i} \neq \hat{\vec{y}}_i \big\rrbracket
\enspace ,
\end{equation}
where $\llbracket \cdot \rrbracket$ is the indicator function.
Although it is frequently used in the literature, one could criticize it for penalizing mistakes in the prediction overly stringent.
In particular, there is no difference between almost correct and completely wrong predictions since the label sets need to match completely.
Consequently, observed values for the subset 0/1 loss are usually relatively high and often close to 1.

The \textit{Hamming} loss represents another extreme as a generalization of the error rate, counting the number of entries in $\widehat{Y}$ deviating from the expected values in $Y$ and dividing by the total number of entries.
\begin{equation}\label{eq:subset01}
\mathcal{L}_{H} (Y_\text{test}, \widehat{Y}) \defeq \dfrac{1}{S} \sum_{i=1}^S \frac{1}{m} \sum_{j=1}^m  \, \big\llbracket y_{i,j} \neq \hat{y}_{i,j} \big\rrbracket \enspace .
\end{equation}
Note that the Hamming loss does not fall into one of the three categories exclusively but can be seen as a representative of all three categories.
Nevertheless, we write it here in the form of an instance-wise loss function.
As the subset 0/1 loss, the Hamming loss behaves rather extreme, and its usefulness for real-world applications might appear debatable.
This is because, in practice, the number of irrelevant labels often outnumbers the number of relevant labels, i.e., the label matrix is usually very sparse, which in turn results in a very low loss.
Hence, even a constant classifier always predicting all labels to be irrelevant has a Hamming loss close to 0.

To find a compromise between these two extremes, a generalized family of instance-wise loss functions, building on so-called non-additive measures, can be considered to interpolate between Hamming and subset 0/1 loss \cite{hullermeier2020flexible}.
For this purpose, this interpolation considers all possible subsets of labels with a certain size and evaluates the average subset 0/1 loss for all the subsets.
Hamming and subset 0/1 loss can be derived as special cases of this generalized family of loss functions if all subsets of size $1$ or $|\mathbb{L}|$ are considered, respectively.
Obviously, the latter considers the predicted and expected label sets as a whole.
The former, in turn, considers $|\mathbb{L}|$ many subsets of size $1$, thus, considering each label of an instance individually and counting the errors.
In between, more emphasis is put on getting subsets of labels with a specific size correctly and thereby allows for interpolating between Hamming and the subset 0/1 loss.

As already mentioned above, in practice, one can frequently observe that the irrelevant labels outnumber the relevant ones by a large margin.
To address this problem of class imbalance, the F1-measure can be considered as it is defined as the harmonic mean of precision and recall.
Note that the F1-measure is not a loss function but rather a measure of accuracy and thus to be maximized\footnote{Obviously, since F1 ranges between 0 and 1, it can be translated into a loss function by taking $1-\text{F1}$.}.
It has been adapted to the MLC setting in the spirit of all three categories, i.e., as an instance-wise, label-wise, and micro loss function which are defined as follows:
\begin{equation}\label{eq:instance-fmeasure}
    \text{F1}_I(Y_\text{test}, \widehat{Y}) \defeq \dfrac{1}{S}\,\, \sum_{i=1}^{S} \dfrac{2 \sum_{j=1}^m y_{i,j}\, \hat{y}_{i,j}}{\sum_{j=1}^{m}(y_{i,j}+ \hat{y}_{i,j})}
\end{equation}
\begin{equation}\label{eq:label-fmeasure}
    \text{F1}_L(Y_\text{test}, \widehat{Y}) \defeq \dfrac{1}{m}\,\, \sum_{j=1}^{m} \dfrac{2 \sum_{i=1}^S y_{i,j}\, \hat{y}_{i,j}}{\sum_{i=1}^{S}(y_{i,j}+\hat{y}_{i,j})}
\end{equation}
\begin{equation}\label{eq:micro-fmeasure}
    \text{F1}_\mu(Y_\text{test}, \widehat{Y}) \defeq \dfrac{2 \sum_{j=1}^{m}\sum_{i=1}^S y_{i,j}\, \hat{y}_{i,j}}{\sum_{j=1}^{m}\sum_{i=1}^{S}(y_{i,j}+\hat{y}_{i,j})}
\end{equation}
While the instance-wise F1-measure accounts for the imbalance between relevant and irrelevant labels for every observation, the label-wise F1-measure considers each label column individually and accounts for imbalance in terms of the respective label being rarely relevant or irrelevant. F1$_\mu$ does not distinguish between labels or instances and more generally emphasizes predicting the relevant labels correctly.
Nevertheless, good performance in terms of the F1-measure requires both a high true positive rate, i.e., predicting relevant labels correctly, and a high true negative rate, meaning to predict irrelevant labels correctly.
Thus, as opposed to $\mathcal{L}_H$, the constant ``always positive'' or ``always negative'' predictor will be assigned the worst performance.


As can already be seen from the example of the constant predictor, the perception regarding the quality of predictions depends significantly on the considered performance measure.
More specifically, a classifier performing best in terms of one loss function does not necessarily perform best for another loss function \cite{hullermeier2020flexible}.
Consequently, this also means, in particular, that the choice and configuration of an MLC classifier need to be tailored to the loss function in question.

\subsection{Label Dependence}\label{ssec:mlc-label-dependence}
One of the major themes and, arguably, the driving force in the multi-label classification literature deals with the development of methods that can exploit so-called \textit{label dependence} to improve the generalization performance.
Label dependence refers to a stochastic correlation between labels, such as labels being positively correlated and thus more likely to occur together.

To give an intuition of what is meant by label dependence, consider again the example of Section~\ref{sec:running-example}. Arguably, the class label \texttt{BEACH} is positively correlated with the class label \texttt{SEA} since beaches are usually located by the sea.
The two class labels are positively correlated.
To put it differently, if the label \texttt{BEACH} is positive, i.e., a beach is on a picture, then the label \texttt{SEA} is likely to be positive as well.
Hence, in this case, a multi-label classifier should be able to acknowledge this correlation and be more reluctant to predict \texttt{BEACH} without \texttt{SEA}, whereas none of the two labels, only \texttt{SEA}, and both labels simultaneously seem to be reasonable predictions.

Formally, label dependence can be distinguished into two categories: \textit{conditional} and \textit{marginal (unconditional)} label dependence \cite{dembczynski2012label}.
While the former refers to a dependence between labels conditioned on a particular instance $\vec{x}$, the latter is of a more general type and can be seen as a kind of \textit{expected} label dependence across the entire instance space.

Following the definitions of conditional and marginal label dependence by \citet{dembczynski2012label}, let $\mathcal{Y} = \{ 0,1\}^m$ be a label space with $m$ labels. Furthermore, let $\vec{Z} = (Z_1, \ldots, Z_m)$ denote a random vector of labels for a probability distribution $P( \cdot , \cdot )$ on $\mathcal{X}\times\mathcal{Y}$, as in Section~\ref{ssec:mlc-problem-setting}. In this context, $\vec{y} \in \mathcal{Y}$ is a realization of this random vector $\vec{Z}$.

For a given instance $\vec{x} \in \mathcal{X}$, a random vector of labels $\vec{Z}$ is called conditionally independent, if 
$$
P(\vec{Z} \mid \vec{x}) = \prod_{i=1}^m P(Z_i \mid \vec{x}) \,\, .
$$
With $P(\vec{Z} \mid \vec{x})$ we denote the conditional probability of observing a random vector $\vec{Z}$ given the instance $\vec{x}$.
Hence, if the probability of the random vector $\vec{Z}$ given an instance $\vec{x}$ is equal to the product of the probabilities of the individual labels given instance $\vec{x}$, the labels are stochastically independent. Vice versa, this means that if this equality does not hold, we observe a conditional label dependence.

Marginal (unconditional) label dependence can be defined similarly. A random vector is called \textit{marginally independent}, if
$$
P(\vec{Z}) = \prod_{i=1}^m P(Z_i) \,\, .
$$
Note that for marginal label dependence the probability is \textit{not} conditioned on any instance $\vec{x}$. Likewise, we observe a marginal label dependence if the equation does not hold.

Furthermore, the two definitions are closely related to each other. As already briefly mentioned before, marginal label dependence can be seen as a kind of \textit{expected} label dependence. This is because we obtain the definition of marginal label dependence by averaging conditional label dependence over the instance space $\mathcal{X}$:
$$
P(\vec{Z}) = \int_\mathcal{X} P(\vec{Z} \mid \vec{x})\, d\mu(\vec{x}) \,\, ,
$$
where $\mu(\cdot)$ denotes the probability distribution on $\mathcal{X}$ according to the joint probability distribution $P(\cdot,\cdot)$. However, despite this relation, \citet{dembczynski2012label} prove that neither dependence implies the other.

While multi-label classifiers can be more or less suitable for either type and may exploit this information, it is not necessarily important to do so for risk minimization.
Depending on the target loss function, an optimal prediction might or might not require taking label dependence into account.
For example, if the loss function is label-wise decomposable, i.e., the loss is first computed over all observations for each label individually and then aggregated, from a theoretical perspective, there is no need to consider label dependencies \cite{dembczynski2012label}.
Instead, it suffices to know (learn) the marginal distributions for the corresponding labels in order to make a risk-minimizing prediction.
Examples of such loss functions are the Hamming loss and the label-wise F1-measure.
Conversely, exploiting label dependence may indeed be advantageous in terms of generalization performance or even necessary to produce risk-minimizing predictions, especially when considering instance-wise loss functions such as the subset 0/1 loss.  

To illustrate the impact of the target loss function on the optimality of a prediction, we consider a simple example, based on our running example of Section~\ref{sec:running-example}.
Let $\mathbb{L} = \{ \texttt{BEACH}, \texttt{FOREST}, \texttt{MOUNTAIN}, \texttt{SEA}\}$ denote the label space of the $m=4$ labels and $\mathcal{Y} = \{0,1\}^4$ correspondingly.
Furthermore, given an observation $\vec{x}$, let the (conditional) ground-truth distribution on $\mathcal{Y}$ be as follows:
\begin{center}
\begin{tabular}{llllc}
\toprule
\texttt{BEACH} & \texttt{FOREST} & \texttt{MOUNTAIN} & \texttt{SEA} & $P( \vec{y} \given \vec{x})$ \\
\midrule
$0$ & $0$ & $0$ & $0$ & $\nicefrac{3}{12}$ \\
$0$ & $1$ & $1$ & $1$ & $\nicefrac{1}{12}$ \\
$1$ & $0$ & $1$ & $1$ & $\nicefrac{2}{12}$ \\
$1$ & $1$ & $0$ & $1$ & $\nicefrac{2}{12}$ \\
$1$ & $1$ & $1$ & $0$ & $\nicefrac{2}{12}$ \\
$1$ & $1$ & $1$ & $1$ & $\nicefrac{2}{12}$ \\
\bottomrule
\end{tabular}
\end{center}
Any other label combination is assigned a probability of 0 and the label dependence is captured via $P(\vec{y} | \vec{x})$.
Note that label dependence is encoded here in terms of the probabilities of the respective label sets, i.e., given instance $\vec{x}$, the most probable label set is $(0,0,0,0)$ with a ground truth probability of $\nicefrac{3}{12}$. However, a 1 is observed with a ground truth probability of $\nicefrac{8}{12}$ for \texttt{BEACH} and $\nicefrac{7}{12}$ for all the remaining labels.

In this example, one can easily see that the Bayes-optimal prediction, which minimizes the loss in expectation, for subset 0/1 loss is $\vec{h}(\vec{x}) = (0,0,0,0)$, i.e., the set of relevant labels is empty ($L_{\vec{x}} = \emptyset$).
However, the Bayes-optimal prediction with respect to the Hamming loss is obtained through $\vec{h}(\vec{x}) = (1,1,1,1)$, i.e., all class labels $L_{\vec{x}} = \{ \texttt{BEACH}, \texttt{FOREST}, \texttt{MOUNTAIN}, \texttt{SEA} \}$ are relevant. Hence, for minimizing the risk in terms of the subset 0/1 loss, it is crucial to consider the joint mode of the distribution and thus the label set $(0,0,0,0)$ as a whole. On the contrary, for an optimal prediction with respect to the Hamming loss, we examine each label individually and thereby ignore the relevance of other labels.
In extreme cases, as in this example, the optimal predictions can even be orthogonal to each other.

As demonstrated in this small example, optimal predictions are not only dependent on the properties of the data, such as the presence of label dependence, but also on the loss function to be minimized.
In particular, this also makes it clear that an MLC algorithm needs to be tailored to both the data -- and therewith the type of label dependence -- and the target loss function.

\subsection{Multi-Label Classifiers}\label{ssec:mlc-classifiers}
Based on the plethora of well-studied SLC algorithms, multi-label classification algorithms have been developed by either \textit{transforming} the original MLC problem into a (set of) single-label classification problem(s) and applying SLC algorithms to the resulting problems or by adapting existing algorithms to the specifics of the MLC problem \cite{DBLP:journals/jdwm/TsoumakasK07}.
More precisely, the algorithms need to be adapted to predict label sets instead of single labels.
In the following, we give a brief overview of both algorithm adaptation methods and problem transformation methods.

\textbf{Problem transformation} algorithms compile the given MLC problem into one or several SLC problems.
Formally, they perform a reduction from the original problem to SLC such that the resulting problems can be dealt with by already known methods, e.g., decision trees, SVMs, or logistic regression.
The most straightforward problem transformation algorithms arguably are the label power set (LP) classifier \cite{boutell2004learning,diplaris2005protein} and binary relevance (BR) learning \cite{DBLP:journals/fcsc/ZhangLLG18}.

LP casts the MLC problem as a single multi-class classification problem, treating each label combination as a distinct class.
Although being conceptually simple, the approach comes with a few technical issues and limitations.
First, the number of classes grows exponentially in the number of labels $m$, and thus, quickly reaches dimensions that render the learning problem unfeasible.
Second, label combinations may be considered as classes that do not and will never actually occur in the data.
Third, explicit information about structures over and dependencies between labels is lost since each label combination is represented by a simple class label. Hence, LP can learn about and account for label dependencies implicitly but not exploit such properties directly.

BR handles the problem transformation quite differently by performing a decomposition into a binary classification problem for each label. Each binary classification problem aims to predict whether the respective label is relevant or not.
Again, due to the implicit independence assumption made about the labels, i.e., that the relevance of each label can be predicted independently of the other labels, interactions or dependencies between labels are ignored in BR as well.
In fact, BR is often criticized for this independence assumption, but for label-wise performance measures, such as Hamming loss or the label-wise F1-measure, it can be theoretically shown that BR is indeed optimal \cite{dembczynski2012label,luaces2012binary}.
Also, the flexibility resulting from the independence assumption can be leveraged to select and configure the SLC methods for tackling the induced binary classification tasks for each label individually.
As shown in Chapter~\ref{ch:libre}, this individualization can be done efficiently, and it benefits the generalization performance.
Nonetheless, the idea of exploiting such label dependencies in order to improve generalization performance is the primary motivation for various methods developed to tackle the MLC problem.

Taking BR as a point of departure, various more sophisticated methods have been developed over time \cite{DBLP:journals/fcsc/ZhangLLG18,read2021classifier}.
A more sophisticated, very powerful, and probably the most prominent problem transformation technique is called \textit{classifier chains} \cite{DBLP:journals/ml/ReadPHF11,read2021classifier}. In classifier chains, a chain of binary classifiers is trained in a sequence such that each classifier predicts the relevance for a specific label but taking into account the relevance of the labels handled by previous classifiers. 
While the relevance of the previous labels is given in the data when training a classifier chain, this information is not available during prediction.
Therefore, the values for the attributes are provided using the predictions of the previous classifiers.
However, a preceding classifier's prediction about the relevance of the respective label might be wrong.
In this case, the error propagates through the chain and affects subsequent predictions, resulting in a kind of attribute noise \cite{DBLP:conf/gfkl/SengeCH12}.
Another issue is that the order of labels within a classifier chain plays an essential role in the generalization performance and has been the subject of optimization in various studies \cite{read2021classifier}.
To overcome this issue, ensembles of classifier chains can be built to compensate for sub-optimal orders of labels \cite{DBLP:journals/ml/ReadPHF11}.

Returning to the image tagging example (cf.~Section~\ref{sec:running-example}), for instance, a classifier chain may first predict the presence of \texttt{BEACH} based on the properties of the picture.
The prediction for \texttt{SEA} could then be conditioned on the properties of the picture \textit{and} the (predicted) presence or absence of the class label \texttt{BEACH}.
In this way, classifier chains may capture a potential dependence between class labels, at least to some extent.
However, the ground truth label for \texttt{BEACH} is only available during training, and during prediction time, it is replaced by a prediction for that label. Hence, if the prediction for \texttt{BEACH} is wrong, the error propagates to the prediction of the class label \texttt{SEA}.

\textbf{Algorithm adaptation} is the other prominent way of developing MLC algorithms, taking SLC methods as a point of departure and adapting models and/or learning algorithms to the MLC setting.
To give a brief impression of algorithm adaptation methods, we present a few such methods in the following. A more detailed overview can be found in \cite{bogatinovski2021comprehensive}.
For example, neural networks can be adapted to multi-label classification quite easily by adding an output neuron for each label and integrating the training algorithm with an MLC loss function \cite{zhang2009m}. Moreover, decision trees have been transferred to the MLC setting by adapting the split criterion \cite{clare2001knowledge} or employing predictive clustering \cite{DBLP:conf/ecml/KocevVSD07}.
A more recent approach, called BOOMER \cite{rapp2020learning}, is a multi-label classification rule learner, leveraging boosting techniques to achieve state-of-the-art generalization performance while providing interpretable models.

\subsection{Configuration of Multi-Label Classifiers}\label{ssec:mlc-configuration}

To achieve the best possible generalization performance, the hyper-parameters of multi-label classifiers need to be tailored to the given data and the loss function in question, as we have already detailed above.
Depending on whether the classifier was designed via algorithm adaptation or a problem transformation, optimizing the hyper-parameters is differently challenging.
While in the former case, only the hyper-parameters directly exposed by the respective multi-label classifier are subject to optimization, in the case of problem transformation, the optimization of hyper-parameters is usually more complex.
Compiling the original MLC problem into a (set of) SLC problem(s), problem transformation methods can be seen as meta-learning methods which are configured with a base learner, i.e., an SLC method, to tackle the resulting SLC problems.
On the one hand, this degree of freedom allows a better fit of the algorithm to the given data and loss function.
On the other hand, for configuring such multi-label classifiers, a decision needs to be made regarding the choice of the base learner and the hyper-parameters that the chosen base learner may additionally expose.

\begin{figure}[t]
    \centering
    \includegraphics{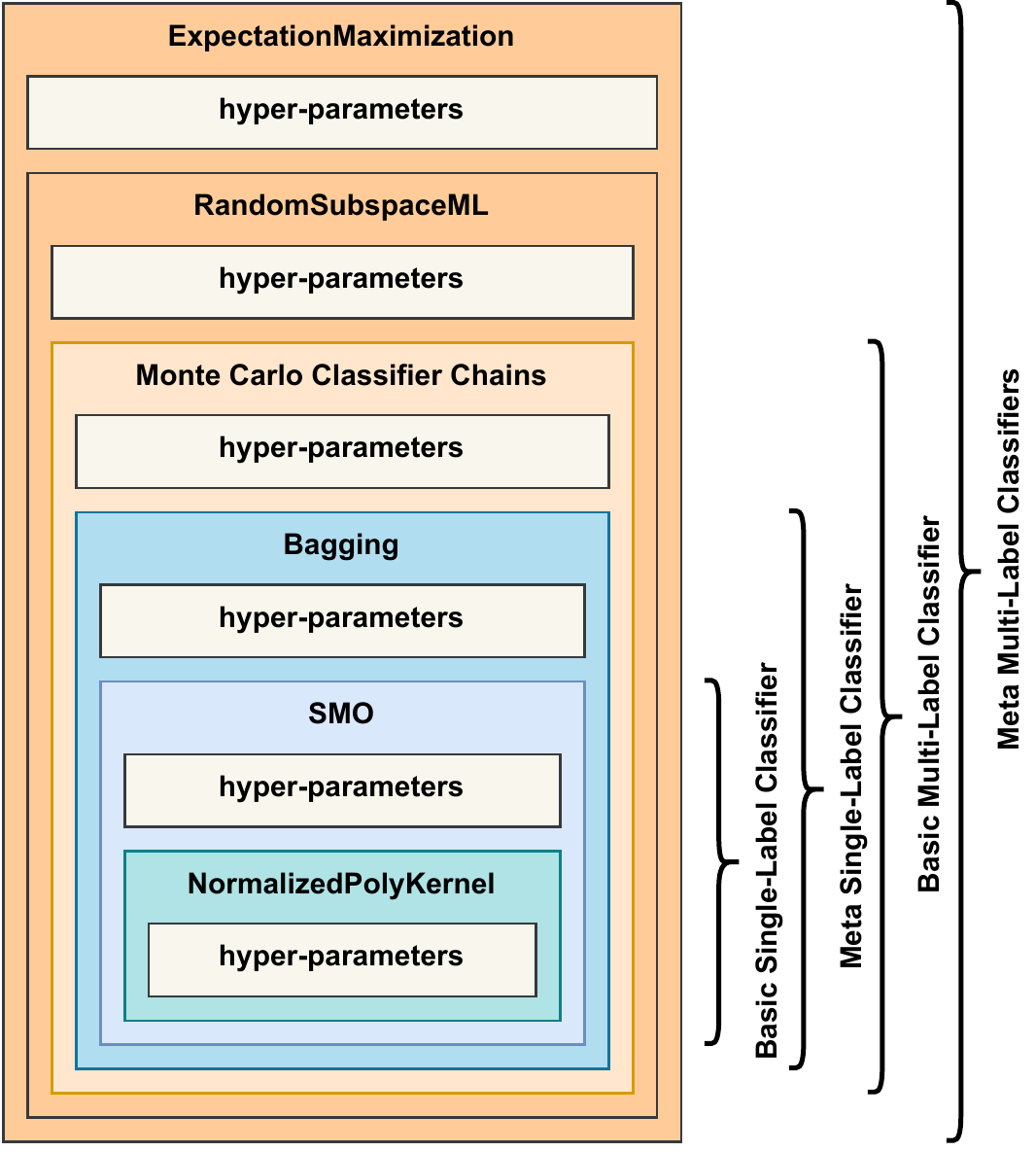}
    \caption{Exemplary illustration of a problem transformation multi-label classifier.}
    \label{fig:mlc-example-pt-method}
\end{figure}

In Figure~\ref{fig:mlc-example-pt-method}, the structure of an exemplary multi-label classification method is illustrated.
The figure displays a multi-label classifier consisting of 5 layers of learning algorithms and a kernel algorithm for the support vector machine (SMO, short for \underline{S}equential \underline{M}inimal \underline{O}ptimization). The outer two layers comprise meta-algorithms for multi-label classification named \texttt{ExpectationMaximization} and \texttt{RandomSubspaceML}. The actual multi-label classifier that is configured as the base learner for the latter meta multi-label classifier is set to be \texttt{Monte Carlo Classifier Chains}, which are in turn configured with a single-label classifier as a base learner. In this example, Bagging is chosen as a meta single-label classifier, which uses \texttt{SMO} as a base learner. Finally, the SMO algorithm is configured with \texttt{NormalizedPolyKernel}. While in this example, the decisions regarding the respective algorithms and base learners are already made, each of the algorithms exposes hyper-parameters that need to be tuned.

\begin{figure}[t]
    \centering
    \includegraphics[width=.48\textwidth]{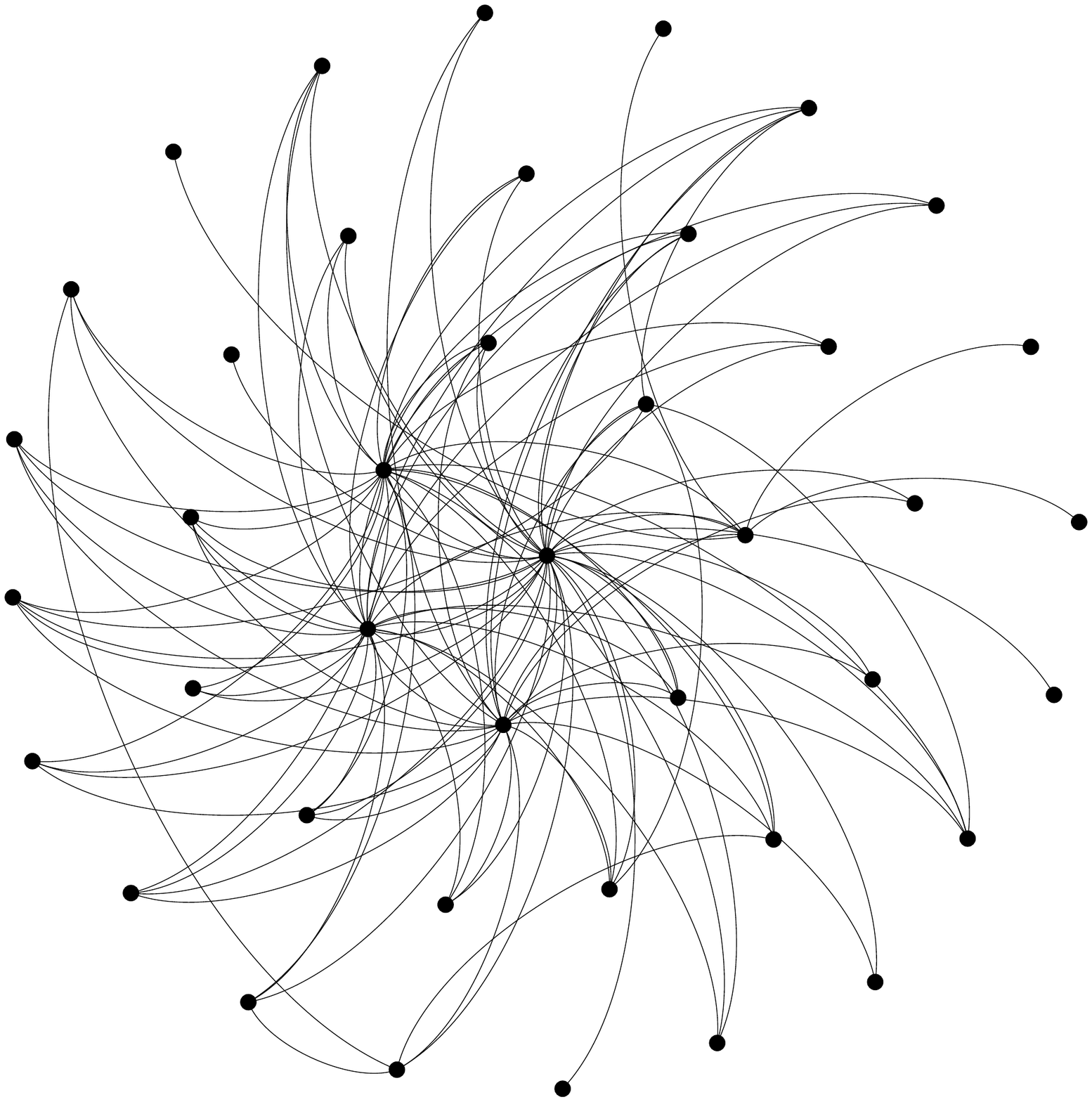}
    \includegraphics[width=.48\textwidth]{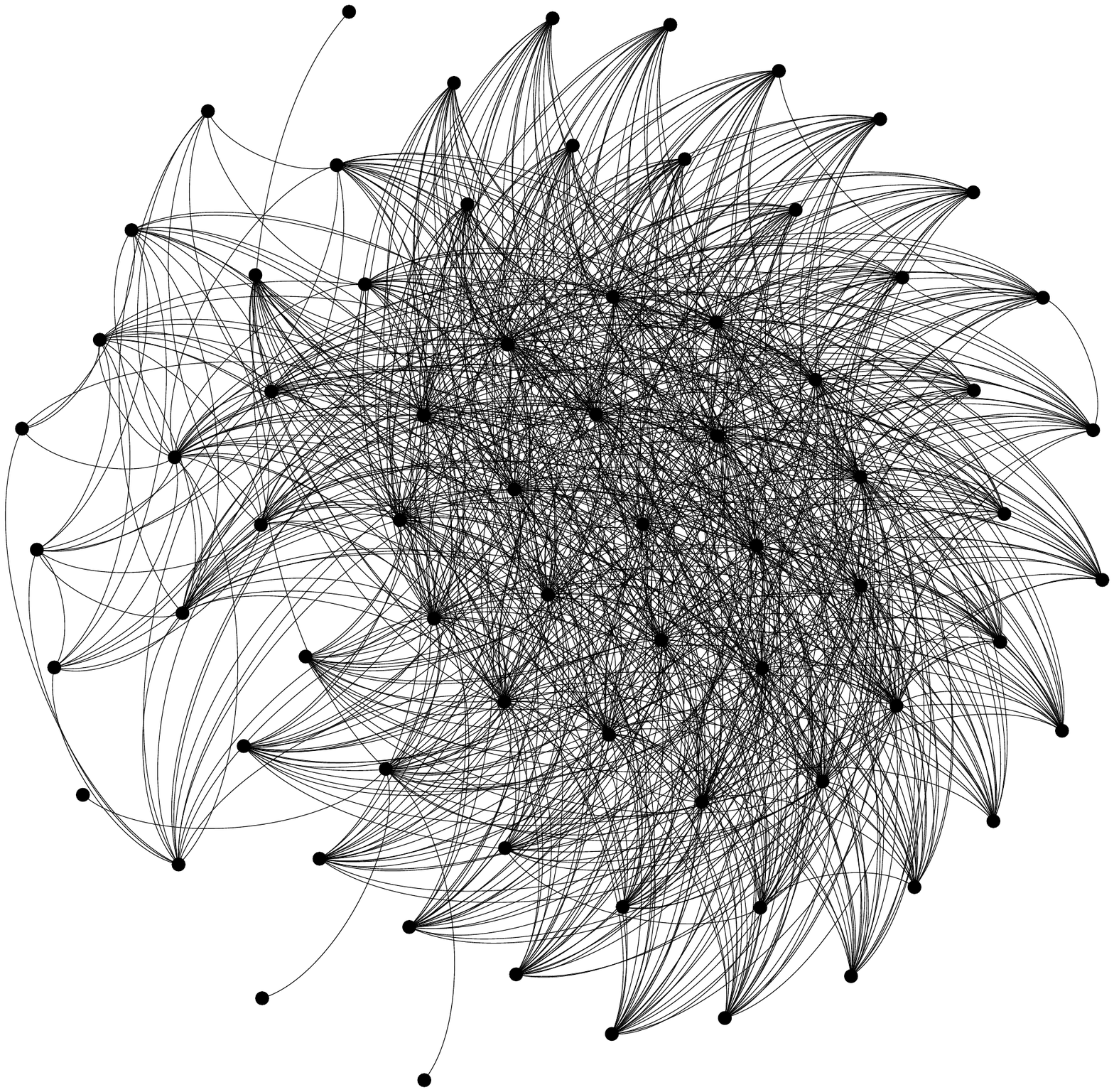}
    \caption{Illustration of the search spaces of ML-Plan for MLC \cite{wever2019automating} only considering the choice of algorithms for single-label classification (\textbf{left}) and multi-label classification (\textbf{right}) as directed acyclic graphs. The curvature of an edge in a clock-wise direction represents a directed edge from a parent to a child node. While each node represents an algorithm contained in the search space, the interpretation of an edge is that the algorithm represented by the parent node has a hyper-parameter that can be configured with the algorithm represented by the child node.}
    \label{fig:mlc-dag-searchspace}
\end{figure}

As can be seen from this single example, the configuration of multi-label classifiers is quite complex, and many decisions need to be made. 
In Chapter~\ref{ch:tpami1}, we deal with the configuration of multi-label classifiers considering a pool of roughly 70 algorithms, exposing a total of more than 170 hyper-parameters (without counting the choice of base learners), whereas a complete multi-label classifier may expose up to 25 hyper-parameters to be configured.
To give an intuition about how much the search space, limited only to the discrete (recursive) choice of algorithms, grows for multi-label classification as compared to the search space for single-label classification, Figure~\ref{fig:mlc-dag-searchspace} shows the respective search spaces as directed acyclic graphs (DAGs).
Nodes of the DAG represent algorithms, and edges indicate a relation between two algorithms in the sense that the child node can be configured as the base algorithm of the parent node.
The direction of an edge can be read from its curvature.
When following an edge in a clockwise direction, one traverses the DAG from a parent node to a child node and vice versa.

While the graph on the left-hand side still shows a clear structure and seems quite manageable, the graph on the right side is much denser.
Algorithms (represented by nodes) and relationships between those algorithms, represented by edges, are difficult to recognize and can only be glimpsed.
More precisely, the search space not only contains more algorithms in total, i.e., 70 compared to 30, but also offers more possibilities to combine algorithms and configure them as base algorithms.
Clearly, selecting and configuring a multi-label classifier is a demanding and cumbersome task, even for experts.
This is all the more true for non-experts.

Going beyond existing hyper-parameters and the configuration of base learners, one may, of course, also think of adding more hyper-parameters to multi-label classifiers, thereby increasing the degrees of freedom and improving their generalization performance.
For example, in \cite{nam2019learning} the permutation of base learners in classifier chains is found to have a practical impact on the performance.
Improving the generalization performance by optimizing this permutation has been the subject of various studies \cite{read2021classifier}.
Moreover, rather than committing to a single base algorithm used for all the labels, a base algorithm could be specifically selected for each label.
For the case of BR, in Chapter~\ref{ch:libre}, we demonstrate that a label-wise configuration of base algorithms for BR may indeed prove beneficial.

While this flexibility can positively affect performance, it obviously also increases the complexity of the configuration space by several orders of magnitude.
Consequently, the configuration process becomes even more complex and increases the need for an automated solution to this problem.

\chapter{ML-Plan: Automated Machine Learning via Hierarchical Planning}\label{ch:mlplan}
\textbf{Declaration of the specific contributions of the author}
The implementation of ML-Plan was initially done by Felix Mohr, and later on, refined by Felix Mohr and the author. The experiments for the WEKA backend and the scikit-learn backend have been conducted by Felix Mohr and the author, respectively. Furthermore, the author described and summarized the results of the experiments in the paper, whereas Felix Mohr wrote the remaining sections of the publications. The entire paper was revised repeatedly by all the authors.

\textbf{Reference:} \fullcite{DBLP:journals/ml/MohrWH18}
\cleardoublepage
\ifthenelse{\boolean{incpapers}}{\includepdf[pagecommand={\thispagestyle{plain}},pages=-]{papers/mlplan}}{}

\chapter{ML-Plan for Unlimited-Length Machine Learning Pipelines}\label{ch:automl1}
\textbf{Declaration of the specific contributions of the author} The publication \textit{ML-Plan for Unlimited-Length Machine Learning Pipelines} is based on the idea of the author. For the realization, the extension of ML-Plan to build machine learning pipelines of unlimited length was implemented. The text was mainly written by the author, where the description of the original ML-Plan was contributed by Felix Mohr. All the authors revised the entire paper.

\textbf{Reference:} \fullcite{wever2018ml}
\cleardoublepage
\ifthenelse{\boolean{incpapers}}{\includepdf[pagecommand={\thispagestyle{plain}},pages=-]{papers/mlplan-ul}}{}

\chapter{Automating Multi-Label Classification Extending ML-Plan}\label{ch:automl2}
\textbf{Declaration of the specific contributions of the author} 
The publication \textit{Automating Multi-Label Classification Extending ML-Plan} is inspired by the idea of Eyke H{\"u}llermeier to extend ML-Plan to the multi-label classification domain. The realization and the writing of the paper were mainly done by the author. Subsequently, the paper was refined by all the authors.

\textbf{Reference:} \fullcite{wever2019automating}
\cleardoublepage
\ifthenelse{\boolean{incpapers}}{\includepdf[pagecommand={\thispagestyle{plain}},pages=-]{papers/ml2-plan}}{}

\chapter{AutoML for Multi-Label Classification: Overview and Empirical Evaluation}\label{ch:tpami1}
\textbf{Declaration of the specific contributions of the author}
The idea for this publication and the implementation can be attributed to the author. Eyke H{\"u}llermeier helped to describe the configuration of multi-label classifiers. Alexander Tornede contributed the introductory paragraphs of the section about techniques reducing AutoML to hyper-parameter optimization and descriptions of Bayesian optimization, Hyperband, and BOHB. Felix Mohr helped with improving the descriptions of the benchmark. All authors repeatedly revised the paper.

\textbf{Reference:} \fullcite{wevertpami2021}
\cleardoublepage
\ifthenelse{\boolean{incpapers}}{\includepdf[pagecommand={\thispagestyle{plain}},pages=-]{papers/tpami.pdf}}{}

\chapter{LiBRe: Label-Wise Selection of Base Learners in Binary Relevance for Multi-Label Classification}\label{ch:libre}
\textbf{Declaration of the specific contributions of the author} 
The general idea of this publication goes back to Felix Mohr, which was further refined, shaped, and targeted by the author. The implementation, as well as the conduction of the experiments, were done by the author. While the related work section was contributed by Felix Mohr, the remaining parts of the paper were initially written by the author and later revised by all authors.

\textbf{Reference:} \fullcite{libre}
\cleardoublepage
\ifthenelse{\boolean{incpapers}}{\includepdf[pagecommand={\thispagestyle{plain}},pages=-]{papers/libre}}{}

\chapter{Ensembles of Evolved Nested Dichotomies for Classification}\label{ch:ndea}
\textbf{Declaration of the specific contributions of the author} 
The idea for the approach and its implementation can be attributed to the author. The paper was mostly written by the author. Felix Mohr and Eyke H{\"u}llermeier helped in increasing the overall quality of the paper with simplifications of the formalisms and revising the text.

\textbf{Reference:} \fullcite{wever2018ensembles}
\cleardoublepage
\ifthenelse{\boolean{incpapers}}{\includepdf[pagecommand={\thispagestyle{plain}},pages=-]{papers/evond}}{}


\chapter{Predicting Machine Learning Pipeline Runtimes in the Context of Automated Machine Learning}\label{ch:tpami2}
\textbf{Declaration of the specific contributions of the author} The author contributed to the publication \textit{Predicting Machine Learning Pipeline Runtimes in the Context of Automated Machine Learning} by discussing intermediate experiments and results thereof. Furthermore, he implemented a composed runtime predictor for integrating it into ML-Plan. He also contributed to the publication by conducting the experiments to evaluate the benefit of this integration for the AutoML task.

\copyright 2021 IEEE. Reprinted, with permission, from Felix Mohr, Marcel Wever, Alexander Tornede, Eyke Hüllermeier, Predicting Machine Learning Pipeline Runtimes in the Context of Automated Machine Learning, IEEE Transactions on Pattern Analysis and Machine Intelligence, 09/2021.

\textbf{Reference:} \fullcite{mohrtpami2021}
\cleardoublepage
\ifthenelse{\boolean{incpapers}}{\includepdf[pagecommand={\thispagestyle{plain}},pages=-]{papers/runtime_paper.pdf}}{}
\ifthenelse{\boolean{incpapers}}{\includepdf[pagecommand={\thispagestyle{plain}},pages=-]{papers/runtime_supp.pdf}}{}

\chapter{Conclusion and Open Questions}\label{ch:future-direction}

According to the structure of the thesis, we draw conclusions and discuss open questions in two parts. First, we focus on AutoML itself and, in particular, AutoML for multi-label classification. Second, we address the topic of improving the efficiency and efficacy of AutoML systems.

In the first part of the thesis, i.e., in Chapters~\ref{ch:mlplan} to \ref{ch:tpami1}, we have devised a novel AutoML system based on hierarchical task network planning for a natural representation of hierarchical dependencies in the configuration space of machine learning pipelines (cf.~Chapter~\ref{ch:mlplan}).
Furthermore, we demonstrated this search space representation to be flexible enough for dealing with machine learning pipelines that are unlimited in length (cf.~Chapter~\ref{ch:automl1}) as well as for the configuration of multi-label classifiers (cf.~Chapter~\ref{ch:automl2}).
Especially in the multi-label classification scenario, where the configuration space is strongly characterized by hierarchical structures, this type of search space representation combined with a best-first search turns out to be very promising.

However, it has also become clear in Chapter~\ref{ch:tpami1} that the high dimensionality of the underlying AutoML problem for multi-label classification presents significant challenges to well-established optimization approaches such as Bayesian optimization and Hyperband.
The high dimensionality of the CASH problem renders most optimization algorithms more or less ineffective. Although most of these methods still perform significantly better than a random search, a \textit{greedy} best-first search algorithm proves to be the most beneficial in the experiments.

A common assumption made in the AutoML literature concerns the dependencies between decisions, i.e., algorithm choices and values for hyper-parameters, that require a joint consideration of the CASH problem.
More specifically, it is assumed that, for example, fixing a pre-processing algorithm affects the optimal decision for learning algorithms and vice versa.
The same assumption is made for the optimization of hyper-parameters.
While these assumptions are certainly valid in theory, the question still is whether they are not unnecessarily obstructive in tackling the AutoML problem and whether a more pragmatic solution would work better in practice.

In particular, the research question arises whether the complexity that comes with the high dimensionality of the search space can be made more manageable.
For example, via a divide and conquer strategy, the search space could be divided into smaller search spaces in which known optimization algorithms can still operate effectively \cite{tornede2021coevolution}.
Alternatively, one could also try to learn how to make the search space complexity more manageable by means of meta-learning \cite{perrone2019learning}, trading in theoretical optimality.
Hence, an open research question is whether the search space can be safely pruned, i.e., without excluding the optimum, or in a way that the search space still contains a near-optimal solution.
From a pragmatic perspective, the question remains whether an ''unsafe`` pruning still leads to better solutions or competitive solutions being found within a shorter time because of the reduced search space complexity.

Furthermore, it is questionable whether AutoML should really be considered as a black-box optimization problem or not since knowledge about what is being configured is completely ignored. Leveraging experience or expert knowledge from data scientists, about the provided training data, or about the algorithms which are combined into machine learning pipelines is difficult in a black-box optimization setting.
Per definition of black-box optimization, it is assumed that nothing is known about the function itself.
While this simplifies the AutoML problem on a conceptual level, it also deliberately ignores potentially valuable information.
Making this knowledge accessible to black-box optimization algorithms is usually a non-trivial endeavor. In \cite{mohr2021naive} we propose a very naive and easy-to-implement approach to AutoML, which, despite its simplicity, is highly competitive to state-of-the-art AutoML systems. The basic idea of this work is to consider the AutoML process as a modular step-by-step procedure.
Each step focuses on a specific decision, e.g., choosing a basic learning algorithm or choosing a pre-processing algorithm. Thereby, incorporating expert knowledge becomes easier than integrating it into black-box optimization algorithms.
The good performance of this approach suggests that opening the black box seems to be a promising direction.

The second part of this thesis (Chapters~\ref{ch:libre} to \ref{ch:tpami2}) dealt with how to increase the efficacy -- through additional degrees of freedom in the configuration of learners -- and the efficiency of AutoML systems.
In Chapter~\ref{ch:libre} and \ref{ch:ndea} respectively, it turned out that by optimizing the structure of a nested dichotomy or by optimizing the base learners of binary relevance learning for each label individually, the generalization performance can be improved significantly.
Even if this means a considerable additional effort for the configuration of such a learner, because of the extra degrees of freedom, this optimization can be automated and thus integrated with AutoML systems. Furthermore, we presented in Chapter~\ref{ch:tpami2} a meta-learning approach for predicting the runtimes of machine learning pipelines in order to prevent the execution of machine learning pipelines that would take too long.

Obviously, proposing to extend the search space contradicts the previous discussion concerning the problem of the high dimensional AutoML search space. Therefore, an open question is which decisions should be included in this search space.
Related to this question is, first of all, the one about the importance of hyper-parameters of multi-label classification methods. While the choice of the base learner is acknowledged as an important hyper-parameter (cf.~\cite{rivolli2020empirical}, Chapter~\ref{ch:libre}), if not the most crucial hyper-parameter, it is still an open question how important other hyper-parameters of multi-label classifiers are. 

Another open research question is to what extent considering runtime predictions of multi-label classifiers helps to avoid classifiers that require excessive time for evaluation.
Avoiding such candidates would prevent the optimization process from stalling, and one would expect more solution candidates to be explored within a fixed time budget.

\chapter{Epilog -- On-The-Fly Computing for Machine Learning Services}\label{ch:otf-computing}

As already mentioned at the beginning of the thesis, this work is motivated by our work in the collaborative research center 901 with the title ``On-The-Fly Machine Learning'', we elaborate on the vision of on-the-fly computing for machine learning services in the following.

\textit{On-the-fly (OTF) computing} refers to a computing paradigm dealing with the automatic, on-the-fly configuration and provision of customized IT services, which is investigated in the eponymous collaborative research center (CRC) 901\footnote{https://sfb901.upb.de (accessed 2021-04-21)} \cite{happe2013fly,karl2019case}. To this end, the IT services are composed of base services, which are available in worldwide markets, by so-called OTF providers and tailored to the specific needs of a customer.
The research on methods for the configuration and provision of such services is accompanied by the investigation of methods for
\begin{itemize}[noitemsep]
    \item quality assurance,
    \item protection of market participants,
    \item target-oriented development of markets, and
    \item supporting interaction between participants
\end{itemize}
that account for the specific characteristics of such dynamically changing markets, also referred to as OTF markets.

In the following, we provide more details on the participants and the structure of OTF markets in Section~\ref{sec:otf-markets} as well as the use case scenario of on-the-fly machine learning (Section~\ref{sec:otf-ml}), which can be seen as an extension of AutoML, as introduced in Section~\ref{sec:automl}.

\begin{figure}[t]
    \centering
    \includegraphics[width=\textwidth]{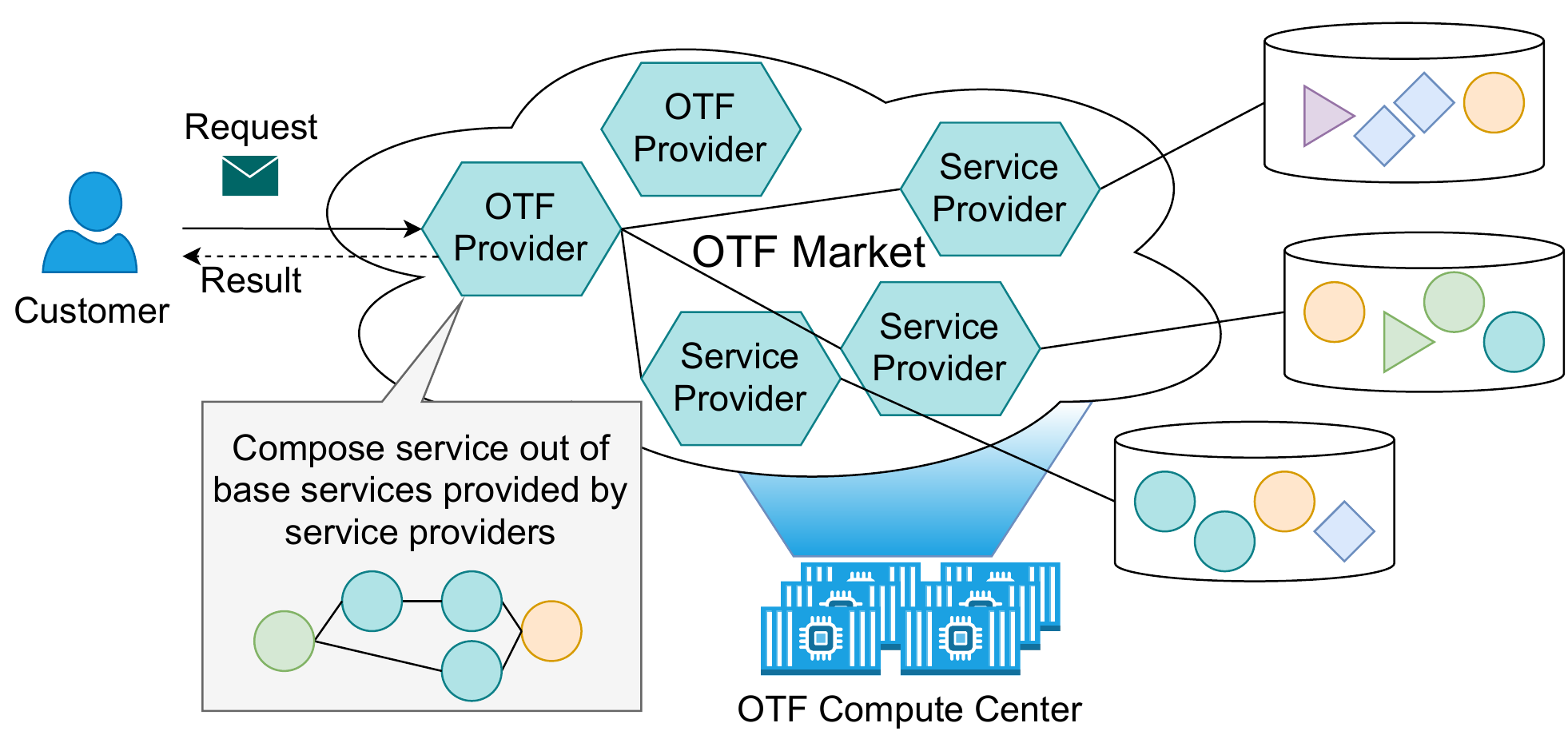}
    \caption{Simplified illustration of an OTF market, involving the roles customer, OTF provider, and service provider. In this illustration, a customer sends a request into the OTF market, which is received and processed by an OTF provider. The OTF provider answers the customer's request by composing a novel service out of base services provided by the service providers to meet the customer's requirements. The result, which might be the composed service itself or the output obtained by executing it, is eventually returned to the customer.}
    \label{fig:otf-market}
\end{figure}

\section{On-The-Fly Markets}\label{sec:otf-markets}
The overall ecosystem of an OTF market comprises many entities ranging from underlying hardware in so-called OTF compute centers to managing entities for setting up markets to (human) participants of the market, which are grouped into roles \cite{jazayeri2017variability}. In the following, we focus on the roles of \textit{customers}, \textit{OTF providers} and \textit{service providers} since those are the ones primarily involved in the process of on-the-fly provisioning software services.
\begin{description}
    \item[Customer] As is implied by the term ``on-the-fly'', in an OTF market, the requested services are provided on demand. More specifically, after a customer sends out a request for a service to the market, the request is processed, and the requested service is composed just then to fulfill the requirements stated in the request. A customer may either be interested in the result of a computation given a specific input, provided with the request, or in the composed service itself. In the latter case, the customer is provided access to the requested service, for example, via a graphical interface or an application programming interface (API).
    \item[OTF Provider] The OTF provider is responsible for receiving, processing, and answering requests. While processing requests means ''understanding`` what is requested by the user, answering the requests involves the automatic composition of the requested service out of base services that are made available in the market by service providers.
    \item[Service Provider] A service provider maintains a repository of base services and provides information about and access to these base services to OTF providers. Obviously, the functionality that can be provided to the user is highly dependent on what base services are provided by service providers.
\end{description}
An illustration of these roles and their interaction is displayed in Figure~\ref{fig:otf-market}.

\section{On-The-Fly Machine Learning}\label{sec:otf-ml}
While OTF computing in general deals with all kinds of IT services, a specific instance of the OTF paradigm solely dealing with machine learning services is referred to as on-the-fly machine learning (OTF-ML) \cite{mohr2018fly,mohr2019automated}.
In this particular scenario, the customer is provided services with machine learning functionality that needs to be tailored to the data in question, i.e., the customer's data for which he or she needs the machine learning functionality.
To this end, the market provides machine learning algorithms as base services, and OTF providers compose these services into service-based machine learning pipelines, where the general concept of a machine learning pipeline remains the same but, instead of algorithms, the services representing the respective algorithms are composed into a machine learning service pipeline \cite{mohr2018wip,mohr2018automated}.

\begin{figure}[t]
    \centering
    \includegraphics[width=\textwidth]{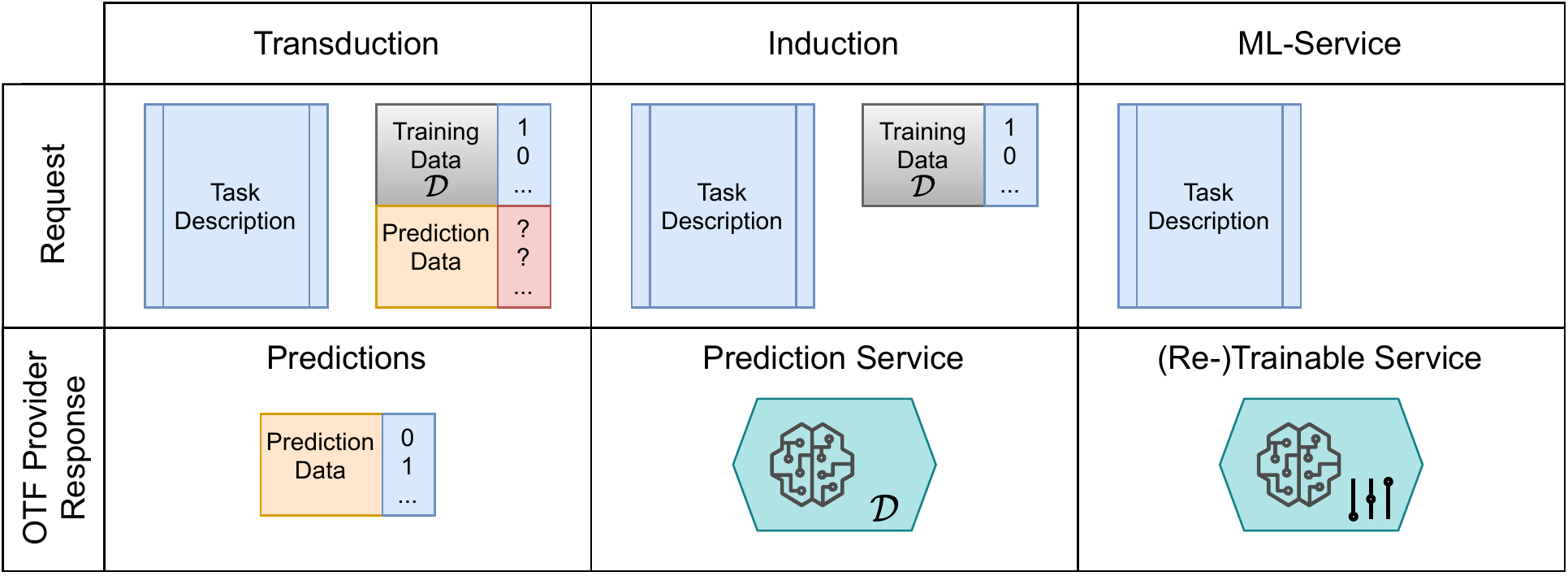}
    \caption{Comparison of different scenarios in on-the-fly machine learning. The scenarios differ in what is contained in the request and what needs to be provided by a customer as well as the desired output, which ranges from one time predictions to a service customized for a provided data set $\mathcal{D}$ that can be repeatedly used for making predictions on new data points to a trainable machine learning service. In the latter scenario, the customer does not provide any data to the OTF provider but trains the service his- or herself.}
    \label{fig:otf-ml}
\end{figure}

Generally speaking, we distinguish three different types of on-the-fly machine learning services that can be requested by a customer \cite{mohr2019automated}:
\begin{description}[noitemsep]
    \item[Transduction] In the transduction scenario, along with a task description, the customer provides a training data set $\mathcal{D}$ together with a set of data points for which the customer is interested in obtaining predictions, referred to as prediction data in Figure~\ref{fig:otf-ml}. The answer to this request consists of the predictions for the given prediction data.
    \item[Induction] In the induction scenario, the customer requests a service that can be used (repeatedly) for labeling new data points. To this end, the customer sends a request containing a task description and training data for which the customer wants to obtain a corresponding machine learning service.
    \item[ML-Service] In this last scenario, the customer only specifies the task the desired machine learning service is meant to accomplish. Without providing data, the task of the OTF provider is thus to provide a customized machine learning service that anyhow performs well for this specific task once data for training is provided.
\end{description}

Deploying AutoML services in an OTF environment offers many advantages. First, the cloud infrastructure offers more flexible and performant computational resources that can be used to achieve a high degree of parallelization for individual AutoML processes, allowing for a faster exploration of the search space.
Furthermore, the abstraction from the platform through services allows for building cross-platform machine learning pipelines, i.e., machine learning pipelines comprising algorithms that are only available for certain platforms.
More specifically, on a service level, it is possible to combine algorithms implemented in C, Python, and Java into a single pipeline.

Besides building machine learning service pipelines from scratch via AutoML, service providers may also offer pre-trained machine learning services specialized in specific tasks, e.g., object recognition for image data. While there might be multiple services offering the same functionality, they may differ in non-functional properties, say, financial costs and accuracy. For example, one service might be very cheap but sometimes very inaccurate, and another service is very accurate on average, but this also comes at a higher cost. Furthermore, other services might be available with non-functional properties ranging between those two extremes.
In \cite{chen2020call,chen2020frugalml}, it is shown that combining such services and deciding for each data point individually which service to use can reduce the overall costs while keeping the quality of predictions competitive to the most accurate single service. Moreover, combining the predictions of multiple services might even result in higher accuracy, as demonstrated in \cite{chen2021frugalmct} within a multi-label classification setting.

{%
\setstretch{1.1}
\renewcommand{\bibfont}{\normalfont\small}
\setlength{\biblabelsep}{0pt}
\setlength{\bibitemsep}{0.5\baselineskip plus 0.5\baselineskip}
\printbibliography[nottype=online,title={References},notkeyword=void]
}
\cleardoublepage
\appendix\cleardoublepage
%

\chapter*{Own Publications}\label{sec:own-publications}
\addcontentsline{toc}{chapter}{Own Publications}

\section*{Journal Articles}

\nocite{wever2020multioracle,merten2019grammatikwandel}
\newrefcontext[sorting=ynt]
\printbibliography[keyword=journal,heading=none]

\section*{Conference Articles}

\nocite{hanselle2020hybrid,tornede2020run2survive,DBLP:conf/pakdd/HanselleTWH21,mohr2018reduction,wever2017active}
\printbibliography[keyword=conference,heading=none]

\section*{Workshop Papers}

\nocite{tornede2020towards,tornede2019algorithm}
\printbibliography[keyword=workshop,heading=none]

\section*{Preprints}

\nocite{merten2021uncertainty,heid2020reliable}
\printbibliography[keyword=preprint,heading=none]

\cleardoublepage

\addcontentsline{toc}{chapter}{\listfigurename}
\listoffigures
\cleardoublepage

%
\pagestyle{empty}
\hfill
\vfill
\pdfbookmark[0]{Colophon}{Colophon}
\section*{Colophon}

This thesis was typeset with \LaTeXe.
It uses the \textit{Clean Thesis} style developed by Ricardo Langner.
The design of the \textit{Clean Thesis} style is inspired by user guide documents from Apple Inc.

Download the \textit{Clean Thesis} style at \url{http://cleanthesis.der-ric.de/}.

\cleardoublepage

%
\pdfbookmark[0]{Declaration}{Declaration}
\chapter*{Declaration}
\label{sec:declaration}
\thispagestyle{empty}

I hereby declare that I have written this dissertation independently and have not used any auxiliary materials other than those indicated. The dissertation has not yet been submitted to any other faculty. I declare that I have not yet unsuccessfully completed a doctoral examination and that I have not been deprived of any doctoral degree that I have already obtained.

\bigskip

\noindent\textit{\thesisUniversityCity, \thesisDate}

\smallskip

\begin{flushright}
	\begin{minipage}{5cm}
		\rule{\textwidth}{1pt}
		\centering\thesisName
	\end{minipage}
\end{flushright}


\clearpage
\newpage
\mbox{}

\end{document}